\def\year{2021}\relax

\documentclass[letterpaper]{article} 
\usepackage{aaai21}  
\usepackage{times}  
\usepackage{helvet} 
\usepackage{courier}  
\usepackage[hyphens]{url}  
\usepackage{graphicx} 
\usepackage{natbib}  
\usepackage{caption} 
\usepackage{booktabs}
\usepackage{amsmath}
\usepackage{amssymb}
\usepackage{bbm}
\usepackage{siunitx}
\usepackage{multirow}
\usepackage{array}
\usepackage{enumitem}

\DeclareMathOperator{\Beta}{Beta}
\DeclareMathOperator{\Bernoulli}{Bernoulli}

\DeclareMathOperator{\diag}{diag}

\DeclareMathOperator{\Cat}{Categorical}
\DeclareMathOperator*{\argmax}{arg\,max}

\newcolumntype{C}[1]{>{\centering\let\newline\\\arraybackslash\hspace{0pt}}m{#1}}

\allowdisplaybreaks

\title{Knowledge-Guided Object Discovery with Acquired Deep Impressions}
\author{
	Jinyang Yuan,
	Bin Li\protect\thanks{Corresponding author},
	Xiangyang Xue \\
}
\affiliations{
	Shanghai Key Laboratory of Intelligent Information Processing, School of Computer Science, Fudan University \\
	\{yuanjinyang, libin, xyxue\}@fudan.edu.cn
}

\begin{document}
	
\maketitle

\begin{abstract}
We present a framework called Acquired Deep Impressions (ADI) which continuously learns knowledge of objects as ``impressions'' for compositional scene understanding. In this framework, the model first acquires knowledge from scene images containing a single object in a supervised manner, and then continues to learn from novel multi-object scene images which may contain objects that have not been seen before without any further supervision, under the guidance of the learned knowledge as humans do. By memorizing impressions of objects into parameters of neural networks and applying the generative replay strategy, the learned knowledge can be reused to generate images with pseudo-annotations and in turn assist the learning of novel scenes. The proposed ADI framework focuses on the acquisition and utilization of knowledge, and is complementary to existing deep generative models proposed for compositional scene representation. We adapt a base model to make it fall within the ADI framework and conduct experiments on two types of datasets. Empirical results suggest that the proposed framework is able to effectively utilize the acquired impressions and improve the scene decomposition performance.
\end{abstract}

\section{Introduction}

\begin{figure}[t]
	\centering
	\includegraphics[width=0.99\columnwidth]{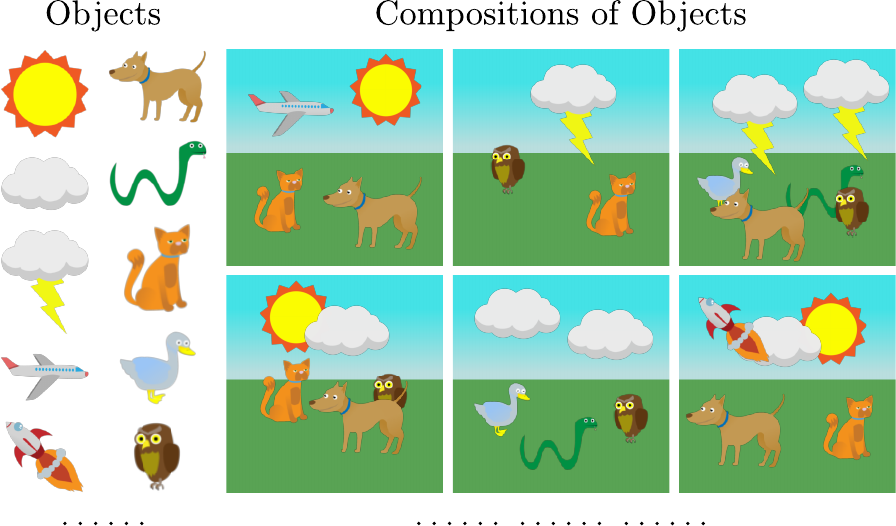}
	\caption{Scenes are diverse compositions of various objects.}
	\label{fig:combination}
\end{figure}

The world is complex not only in the great variations of objects that constitute it, but also in the diverse compositions of these objects. Figure \ref{fig:combination} presents some toy examples of scenes constructed from various simple objects. In order to interact with the world efficiently and effectively, humans tend to understand the perceived visual scenes in a compositional and structured way by decomposing the complex scenes into relatively simple objects and organizing these objects structurally \cite{lake2017building}. Regularized by possibly inborn mental laws \cite{koffka2013principles,goldstein2016sensation} and based on the previously learned knowledge of objects, humans can decompose and understand novel visual scenes composed of familiar objects reliably, and build up new knowledge of novel objects efficiently with few examples of visual scenes containing these objects. The newly acquired knowledge is accumulated in the memory and assists the understanding of new scenes in the future.

In recent years, a large body of deep generative models such as variational autoencoders (VAE) \cite{kingma2013auto,rezende2014stochastic} and flow-based generative models \cite{dinh2014nice,dinh2016density,kingma2018glow} have been proposed and provide mechanisms to represent images with latent variables by building nonlinear mappings between images and latent variables with neural networks. The prior distribution of the latent variable along with the neural network characterize the distribution of images and can thus be seen as the learned knowledge of images. By sampling latent variables from the prior distribution and transforming latent variables with the decoding neural network, novel images similar to those used for training can be generated. Most existing VAEs and flow-based models learn a single representation for the whole image, and cannot be used to decompose the scene image into individual objects directly.

\begin{figure}[t]
	\centering
	\includegraphics[width=0.96\columnwidth]{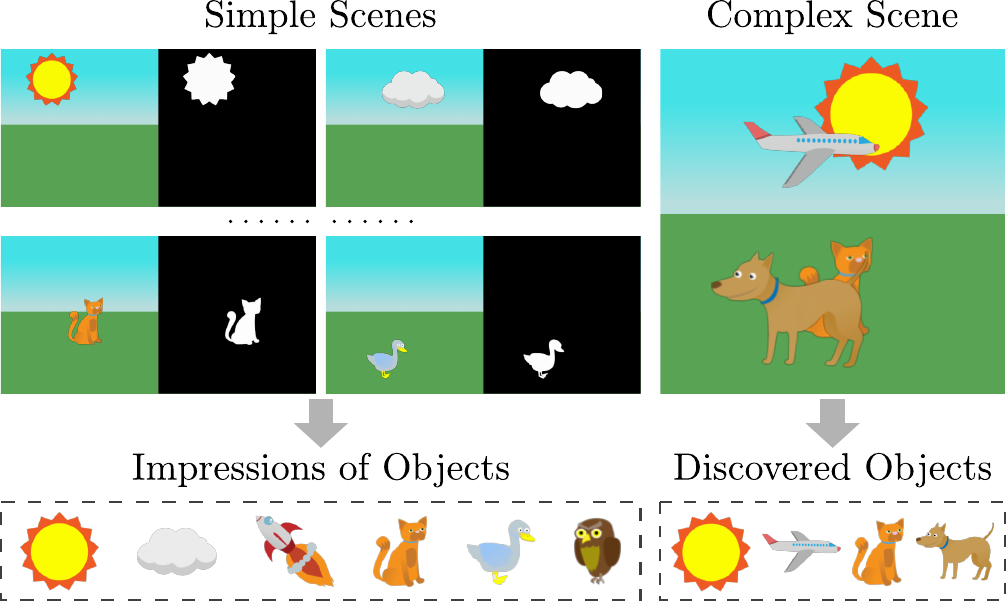}
	\caption{Complete objects in the complex scene can be discovered using impressions learned from simple scenes.}
	\label{fig:knowledge}
\end{figure}

It is intriguing to design human-like machines that are able to compositionally represent scenes with individual objects and continuously learn knowledge of objects which acts as priors to assist the decomposition of novel scenes. In contrast to the aforementioned approaches which treat the full scene as a whole and lack effective mechanisms of incorporating accumulated prior knowledge, such machines lower the complexity to represent visual scenes and facilitate understanding novel visual scenes. As shown in Figure \ref{fig:knowledge}, under the guidance of impressions of objects which are previously acquired in a supervised manner, it is possible to extract complete objects from the perceived visual scene even if the scene contains some novel objects that have not been seen before and occlusions exist.

To decompose visual scenes and achieve better compositionality, a number of deep generative models \cite{huang2015efficient,eslami2016attend,greff2017neural,crawford2019spatially,stelzner2019faster,greff2019multi,yuan2019generative,burgess2019monet,engelcke2019genesis,yang2020learning} which learn a separate representation for each object in the scene in an unsupervised manner, have been proposed. Object-based representations extracted by these methods are fundamental to the acquisition of impressions of objects. However, these works focus mainly on the unsupervised decomposition of scenes instead of the acquisition and utilization of impressions.

In this paper, we propose a framework called Acquired Deep Impressions (ADI) which enables continuously learning knowledge of individual objects constituting the visual scenes as ``impressions'' in a methodical manner. In this framework, visual scenes are modeled as the composition of layers of objects and background using spatial mixture models. Each layer is associated with a latent variable, which can be transformed into appearance and shape of the layer by learnable mappings modeled by \emph{deep} neural networks. The parameters of neural networks are \emph{acquired} from data, and the prior distributions of latent variables together with the learned parameters of neural networks define the \emph{impressions} of objects and background.

Complementary to existing deep generative models proposed for compositional scene representation, the proposed ADI framework provides a mechanism to effectively acquire and utilize knowledge. An existing base model is adapted to fall within the ADI framework, and experiments are conducted with this model on two types of datasets. Extensive empirical results suggest that the impressions previously acquired from relatively simple visual scenes play an important role as prior knowledge for the discovery of objects in complex scenes, and the proposed ADI framework is able to improve the scene decomposition performance by effectively acquiring and utilizing impressions.

\section{Related Work}

In recent years, various spatial mixture models have been proposed for the binding problem \cite{treisman1996binding} and related higher-level tasks which require extracting compositional latent representations of objects. RC \cite{greff2015binding} iteratively reconstructs the visual scene compositionally with a pretrained denoising autoencoder. Tagger \cite{greff2016tagger} utilizes a Ladder Network \cite{rasmus2015semi} to extract both high-level and low-level latent representations which are suitable for different tasks. RTagger \cite{premont2017recurrent} is applicable to sequential data by replacing the Ladder Network used in \cite{greff2016tagger} with the Recurrent Ladder Network. N-EM \cite{greff2017neural} infers latent representations of objects with a recurrent neural network (RNN) based on the Expectation-Maximization (EM) framework. Relational N-EM \cite{van2018relational} tackles the problem of common-sense physical reasoning by modeling relations between objects. SMMLDP \cite{yuan2019spatial} uses neural networks to model mixture weights of the spatial mixture model and infers latent representations based on the N-EM framework. In all these models, priors of latent representations are not defined, and inferences of latent representations are based on the maximum likelihood estimation (MLE). Knowledge of objects are not fully captured by the learned model because there is no natural way to sample latent representations and generate visual scenes similar to the ones used for training.

CST-VAE \cite{huang2015efficient} models visual scenes by layered representations \cite{wang1994representing} and takes occlusions of objects into consideration. AIR \cite{eslami2016attend} is a type of variable-sized VAE that can determine the number of objects in the scene and extract object-based representations. SQAIR \cite{kosiorek2018sequential} extends AIR to videos of moving objects by modeling relations between objects in consecutive frames. SPAIR \cite{crawford2019spatially} combines ideas used in supervised object detection with AIR and is able to handle large and many-objects scenes. SuPAIR \cite{stelzner2019faster} substitutes VAEs in AIR with sum-product networks to increase the speed and robustness of learning. IODINE \cite{greff2019multi} jointly segments and represents objects based on the iterative amortized inference framework \cite{marino2018iterative}. GMIOO \cite{yuan2019generative} models occlusions between objects and is able to determine the number of objects in the scene and segment them simultaneously. MONet \cite{burgess2019monet} proposes a novel recurrent attention network for inferring compositional latent representations. GENESIS \cite{engelcke2019genesis} considers relationships between objects in the generative model and is able to generate more coherent scenes. \cite{yang2020learning} integrates Contextual Information Separation and perceptual cycle-consistency into compositional deep generative model, and is able to perform unsupervised segmentation on manually generated scenes composed of objects with complex textures.

Object-based representations play a fundamental role in the proposed ADI framework. The aforementioned methods learn to extract object-based representations without ground-truth annotations, and can be applied to discover objects in an unsupervised manner. Complementary to these methods, ADI focuses more on the utilization of previously learned knowledge and proposes a learning procedure which could empower these methods by exploiting the acquired impressions to assist the discovery of objects (including those that have not been seen before) in novel scenes. ADI aims to decompose a scene containing possibly multiple objects into object(s) and background, and thus differs from the methods proposed for single-object scenes \cite{singh2019finegan} or the methods that decompose scenes into hierarchical features \cite{zhao2017learning}. Because ADI first acquires impressions from simple scenes in the supervised scenario and then continues to learn from more complex scenes without any further supervision, it is also different from the methods which incorporate pretrained models in the supervised learning pipeline \cite{ulutan2020vsgnet}.

\section{Acquired Deep Impressions}

Representing visual scenes is complex due to the diverse combinations of objects in the scenes. In most existing deep generative models, the whole visual scene is encoded into a single latent representation. To obtain models which generalize well in novel scenes, a great number of data which cover all possible combinations need to be observed during training. Because the combinations of objects are in general extremely complex even if individual objects are simple to model, learning such a type of representation is not very data-efficient. If visual scenes can be decomposed into objects which are represented separately, improved sampling efficiency and generalizability may be achieved. Humans can decompose complex and novel scenes which are composed of familiar objects effectively even if occlusions exist, probably because impressions of complete objects are accumulated from simpler scenes which have been observed previously. Inspired by this phenomenon, we propose a framework called Acquired Deep Impressions (ADI) to facilitate the discovery of objects in novel scenes by learning \emph{impressions} of objects in a methodical manner. The proposed ADI framework is complementary to existing compositional scene representation methods and is able to empower these methods to utilize previously acquired knowledge by modifying them to fall within the ADI framework.

\subsection{Generative Model}

\begin{figure}[t]
	\centering
	\includegraphics[width=0.96\columnwidth]{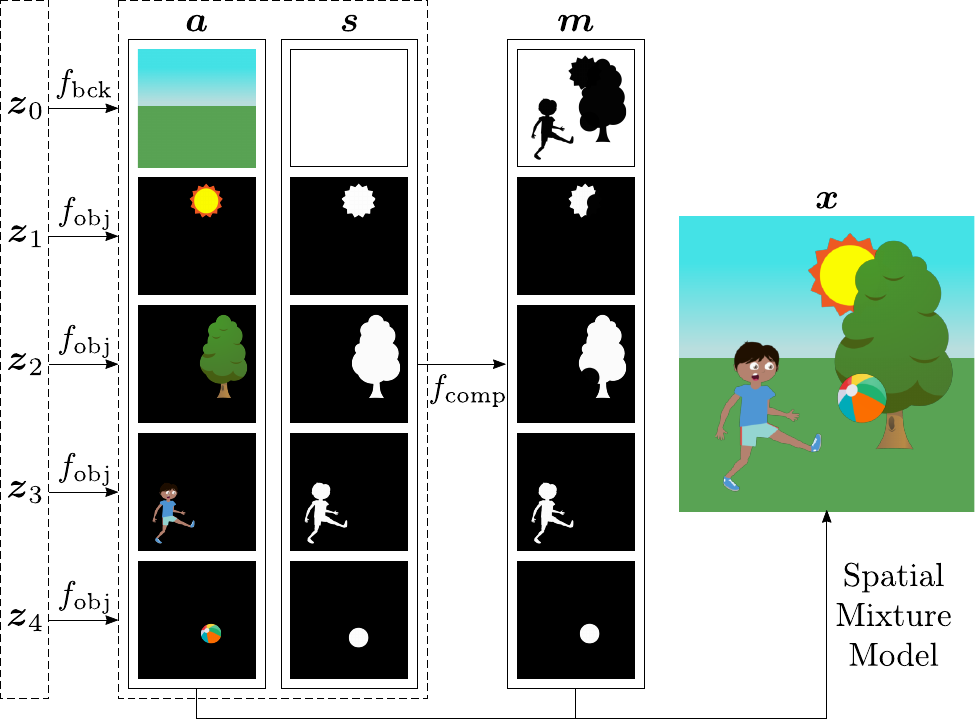}
	\caption{Illustration of the generative process. $\boldsymbol{z}_k$ is the latent variable of background ($k\!=\!0$) or object ($k\!\geq\!1$). $\boldsymbol{a}$, $\boldsymbol{s}$, and $\boldsymbol{m}$ are the collections of appearances, masks of complete shapes (or logits of perceived shapes), and masks of perceived shapes, respectively. $\boldsymbol{x}$ is the generated visual scene.}
	\label{fig:generate}
\end{figure}

In ADI, a visual scene $\boldsymbol{x} \in \mathbb{R}^{N \times C}$ is assumed to be generated by composing layers of objects and background using spatial mixture models. $N$ and $C$ are the respective numbers of pixels and channels in each image. Each layer $k$ is associated with a latent variable $\boldsymbol{z}_k$ that is drawn independently from the prior distribution $p(\boldsymbol{z}_k; \boldsymbol{\theta}_{\text{bck}})$ or $p(\boldsymbol{z}_k; \boldsymbol{\theta}_{\text{obj}})$ for $k \!=\! 0$ (background) or $k \!\geq\! 1$ (objects). The collection of all latent variables $\{\boldsymbol{z}_0, \boldsymbol{z}_1, \dots\}$ is denoted by $\boldsymbol{z}$. Each latent variable $\boldsymbol{z}_k$ is transformed to the appearance $\boldsymbol{a}_k \in \mathbb{R}^{N \times C}$ and the variable containing shape information $\boldsymbol{s}_k \in \mathbb{R}^{N}$ (e.g., mask of complete shape or logit of perceived shape) of the background ($k = 0$) or object ($k \geq 1$), by a learnable mapping $f_{\text{bck}}$ or $f_{\text{obj}}$. A compositing function $f_{\text{comp}}$ (e.g. stick-breaking or softmax) which simultaneously takes all the variables containing shape information $\boldsymbol{s} \!=\! \{\boldsymbol{s}_0, \boldsymbol{s}_1, \dots\}$ as inputs is then applied to transform $\boldsymbol{s}$ into perceived shapes $\boldsymbol{m} \!=\! \{\boldsymbol{m}_0, \boldsymbol{m}_1, \dots\}$ ($\boldsymbol{m}_k \in \mathbb{R}^{N}, \forall k$). Each pixel $\boldsymbol{x}_n$ of the visual scene is assumed to be conditional independent of each other given all the latent variables $\boldsymbol{z}$. Let $l_n$ denote the variable indicating which layer is observed at the $n$th pixel. The perceived shapes $\boldsymbol{m}$ and appearances $\boldsymbol{a}$ of layers are used as mixture weights $p(l_n\!=\!k|\boldsymbol{z})$ and parameters of mixture components $p(\boldsymbol{x}_n|\boldsymbol{z}_{l_n}, l_n\!=\!k)$, respectively. The joint probability of the visual scene $\boldsymbol{x}$ and latent variables $\boldsymbol{z}$ is factorized as $p(\boldsymbol{x}, \boldsymbol{z}) = p(\boldsymbol{z}) p(\boldsymbol{x}|\boldsymbol{z})$, where
\begin{gather}
	p(\boldsymbol{z}) = \prod\nolimits_{k}{p(\boldsymbol{z}_k)} \\
	p(\boldsymbol{x}|\boldsymbol{z}) = \prod\nolimits_{n}{\sum\nolimits_{k}{p(l_n\!=\!k|\boldsymbol{z}) p(\boldsymbol{x}_n|\boldsymbol{z}_{l_n}, l_n\!=\!k)}}
\end{gather}
The generative process of the visual scene is illustrated in Figure \ref{fig:generate}. The detailed expressions are given by
\begin{gather*}
	\begin{aligned}
		\boldsymbol{z}_k & \sim p(\boldsymbol{z}_k; \boldsymbol{\theta}_{\text{bck}}), & \quad \boldsymbol{a}_k, \boldsymbol{s}_k & = f_{\text{bck}}(\boldsymbol{z}_k); & \qquad k & = 0 \\
		\boldsymbol{z}_k & \sim p(\boldsymbol{z}_k; \boldsymbol{\theta}_{\text{obj}}), & \quad \boldsymbol{a}_k, \boldsymbol{s}_k & = f_{\text{obj}}(\boldsymbol{z}_k); & \qquad k & \geq 1
	\end{aligned} \\
	\boldsymbol{m}_0, \boldsymbol{m}_1, \dots = f_{\text{comp}}(\boldsymbol{s}_0, \boldsymbol{s}_1, \dots) \\
	l_n \sim \Cat(m_{0,n}, m_{1,n}, \dots) \\
	\boldsymbol{x}_n \sim p(\boldsymbol{x}_n; \boldsymbol{a}_{l_n, n})
\end{gather*}

This framework is called Acquired Deep Impressions (ADI) because: 1) the learnable mappings from latent variables to objects and background are \emph{acquired} from data; 2) these mappings are modeled by \emph{deep} neural networks; and 3) the priors of latent variables together with the learned parameters of neural networks define the \emph{impressions} of objects and background. The acquired impressions act as strong prior knowledge that assist the model to discover objects and extract latent variables of \emph{complete} objects and background even if the perceived objects and background are incomplete due to occlusions.

\subsection{Variational Inference}

\begin{figure}[t]
	\centering
	\includegraphics[width=0.95\columnwidth]{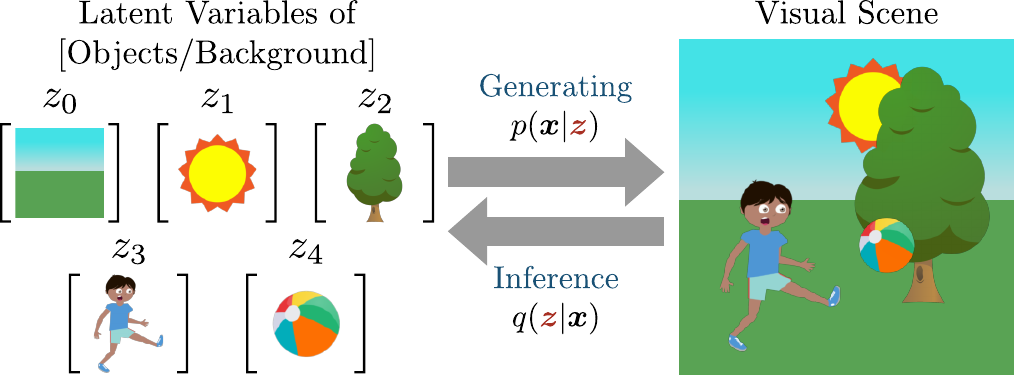}
	\caption{Objects and background are fully characterized by latent variables $\boldsymbol{z}$, and the decomposition of a visual scene is equivalent to the inference of $\boldsymbol{z}$.}
	\label{fig:decomposition}
\end{figure}

As shown in Figure \ref{fig:decomposition}, the layers of objects and background constituting the visual scene are fully characterized by latent variables $\boldsymbol{z}$, and the decomposition of visual scene is equivalent to the inference of latent variables of all the layers. Latent variables are inferred by a neural network $g$ under the variational inference framework. The inference network $g$ takes the visual scene $\boldsymbol{x}$ as inputs, and outputs parameters of the variational distribution $q(\boldsymbol{z}|\boldsymbol{x})$. By training the inference network to approximate the mapping from visual scene to latent variables, the model learns to perform scene decomposition in an amortized manner.

\subsection{Learning Procedure}

Learning to discover objects from scenes is in general difficult without supervision. ADI provides a way to lower the difficulty by dividing the learning into two stages which differ from each other mainly in the data used to train the model and the way the training is conducted:
\begin{itemize}[leftmargin=*]
	\item \textbf{Stage 1:} The model is first trained on scene images containing a single object, under the supervision of manually annotated or automatically generated ground truth of object shapes. Impressions of objects are saved as parameters of the decoding networks that map latent variables $\boldsymbol{z}$ to appearances and shapes of layers as well as the inference networks that output the approximated posterior distributions $q(\boldsymbol{z}|\boldsymbol{x})$. Because only one object may appear in each image, how to handle occlusions of object is not learned in this stage. 
	\item \textbf{Stage 2:} The model then continues to learn from scene images comprising possibly multiple objects, without using any further supervision. Some images contain objects that have not been seen before. The model is expected to learn how to discover \emph{complete} objects, including those not appearing in the first learning stage, even if the observed objects are incomplete due to occlusion. In order to improve the efficiency and effectiveness of learning, impressions acquired in the first learning stage are exploited using the generative replay strategy \cite{shin2017continual}.
\end{itemize}

\begin{figure}[t]
	\centering
	\includegraphics[width=0.95\columnwidth]{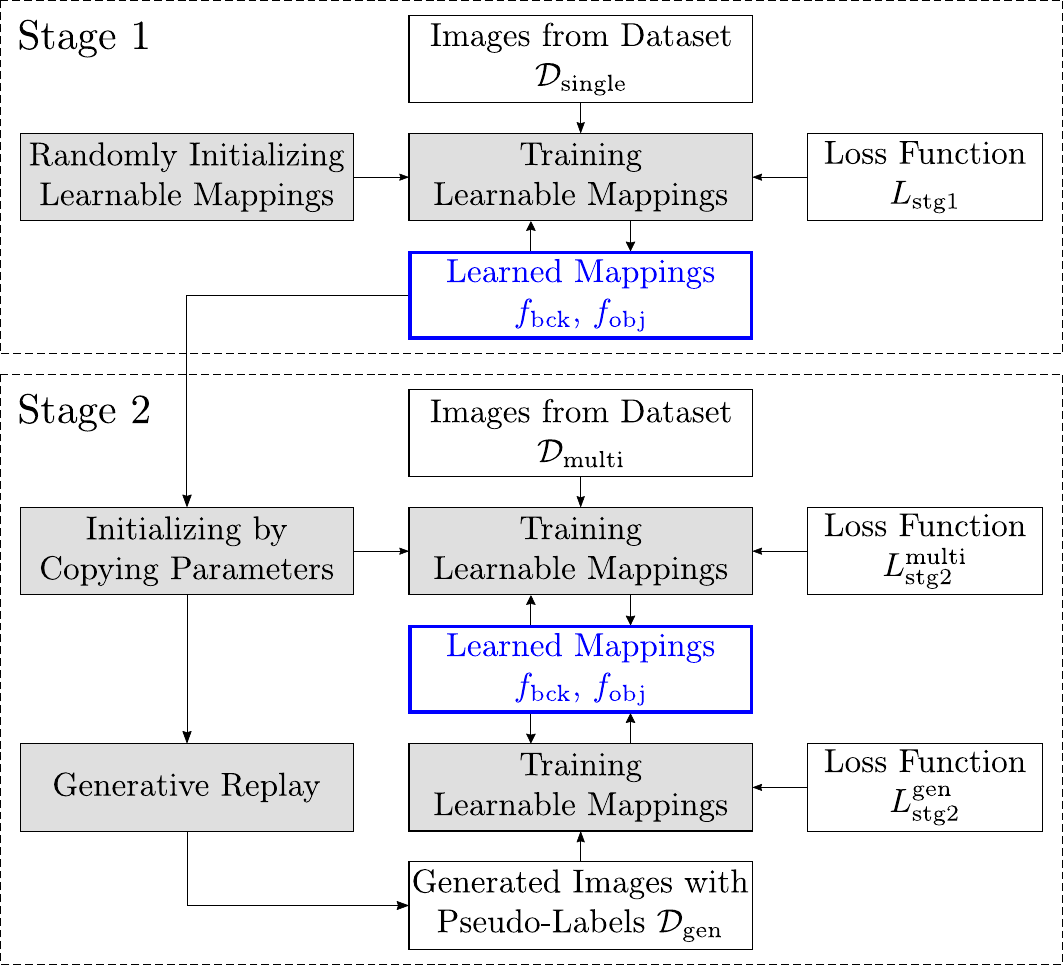}
	\caption{Illustration of the training procedure. In the first learning stage, learnable mappings $f_{\text{bck}}$ and $f_{\text{obj}}$ are iteratively refined by minimizing $L_{\text{stg1}}$ on $\mathcal{D}_{\text{single}}$. In the second learning stage, $f_{\text{bck}}$ and $f_{\text{obj}}$ are iteratively refined by minimizing $L_{\text{stg2}}^{\text{multi}}$ and $L_{\text{stg2}}^{\text{gen}}$ on $\mathcal{D}_{\text{multi}}$ and $\mathcal{D}_{\text{gen}}$ simultaneously.}
	\label{fig:training}
\end{figure}

A commonly used loss function to train variational autoencoders is the negative evidence lower bound (ELBO) $L_{\text{elbo}}$, which can be decomposed into the negative log-likelihood (NLL) term $L_{\text{nll}}$ and the Kullback–Leibler divergence (KLD) term $L_{\text{kld}}$, i.e., $L_{\text{elbo}} = L_{\text{nll}} + L_{\text{kld}}$. The detailed expressions of $L_{\text{nll}}$ and $L_{\text{kld}}$ are
\begin{gather}
	L_{\text{nll}} = -\mathbb{E}_{\boldsymbol{z} \sim q(\boldsymbol{z}|\boldsymbol{x})}\big[\log{p(\boldsymbol{x}|\boldsymbol{z})}\big] \\
	L_{\text{kld}} = \mathbb{E}_{\boldsymbol{z} \sim q(\boldsymbol{z}|\boldsymbol{x})}\big[\log{q(\boldsymbol{z}|\boldsymbol{x})} - \log{p(\boldsymbol{z})}\big]
\end{gather}
Beta-VAE \cite{higgins2017beta} adds an adjustable hyperparameter $\beta \!>\! 1$ as the coefficient of the KLD term in the ELBO to improve the quality of disentanglement, i.e. $L_{\text{elbo}}^{\beta} = L_{\text{nll}} + \beta L_{\text{kld}}$. As long as $\beta$ is no less than $1$, $-L_{\text{elbo}}^{\beta}$ is a valid lower bound of the log-evidence $\log{p(\boldsymbol{x})}$. We design the loss functions based on $L_{\text{elbo}}^{\beta}$, and add extra terms to incorporate supervision and perform generative replay. The illustration of the training procedure is shown in Figure \ref{fig:training}.

In Stage 1, Images $\mathcal{D}_{\text{single}}$ containing a \emph{single} object are used to train the model in a \emph{supervised} manner. The loss function of this learning stage consists of three parts:
\begin{equation}
	L_{\text{stg1}} = L_{\text{nll}} + \beta_{\text{stg1}} L_{\text{kld}} + \alpha_{\text{stg1}} L_{\text{mask}}
	\label{equ:loss_stage1}
\end{equation}
In the above expression, $\beta_{\text{stg1}} \!\geq\! 1$ and $\alpha_{\text{stg1}} \!>\! 0$ are tunable hyperparameters. The first two parts are the NLL and KLD terms, and the third part is the supervision term. The supervision used during the training is the ground truth of mixture weights (perceived shapes) of all layers $\tilde{\boldsymbol{m}}$. The KL divergence between the ground truth $\tilde{\boldsymbol{m}}$ and the estimated mixture weights $\boldsymbol{m}$ are used as the supervision term $L_{\text{mask}}$ in the loss function
\begin{equation}
	L_{\text{mask}} = \sum\nolimits_{k,n}{\tilde{m}_{k,n} (\log{\tilde{m}_{k,n}} - \log{m_{k,n}})}
\end{equation}

In Stage 2, the model continues to learn from \emph{unannotated} images $\mathcal{D}_{\text{multi}}$ containing \emph{multiple} objects. When trained on these data $\mathcal{D}_{\text{multi}}$, the loss function is simply an upper bound of the negative log-evidence:
\begin{equation}
	L_{\text{stg2}}^{\text{multi}} = L_{\text{nll}} + \beta_{\text{stg2}} L_{\text{kld}}
\end{equation}
In order to guide the learning with the help of previously acquired impressions of objects, additional images $\mathcal{D}_{\text{gen}}$ are generated by following the generative process. $f_{\text{bck}}$ and $f_{\text{obj}}$ which are learned in the first learning stage are used to transform the sampled latent variables $\tilde{\boldsymbol{z}}$ into appearances $\tilde{\boldsymbol{a}}$ and variables containing shape information $\tilde{\boldsymbol{s}}$ of all the layers. $\tilde{\boldsymbol{m}} = f_{\text{comp}}(\tilde{\boldsymbol{s}})$ is used as the supervision, and the model is trained on the generated images $\mathcal{D}_{\text{gen}}$ using a loss function differing from Eq. \eqref{equ:loss_stage1} only in the chosen hyperparameters:
\begin{equation}
	L_{\text{stg2}}^{\text{gen}} = L_{\text{nll}} + \beta_{\text{stg2}} L_{\text{kld}} + \alpha_{\text{stg2}} L_{\text{mask}}
\end{equation}

\section{Experiments}

Comprehensive experiments are conducted to validate the effectiveness of the acquired impressions. We adapt an existing method GMIOO \cite{yuan2019generative} proposed for compositional scene representation to make it fall within the ADI framework, and evaluate the adapted model on two types of datasets. \footnote{Code is available at \url{https://github.com/jinyangyuan/acquired-deep-impressions}.} Details of the adaptation and choices of hyperparameters are provided in the supplementary material. Experimental results under different configurations demonstrate that the acquired impressions can greatly improve the discovery of objects in novel scenes.
\footnote{We have also tried to adapt another base model AIR \cite{eslami2016attend}. Experimental results provided in the supplementary material verify that the ADI framework is also effective on AIR.}

\begin{figure}[t]
	\centering
	\includegraphics[width=0.99\columnwidth]{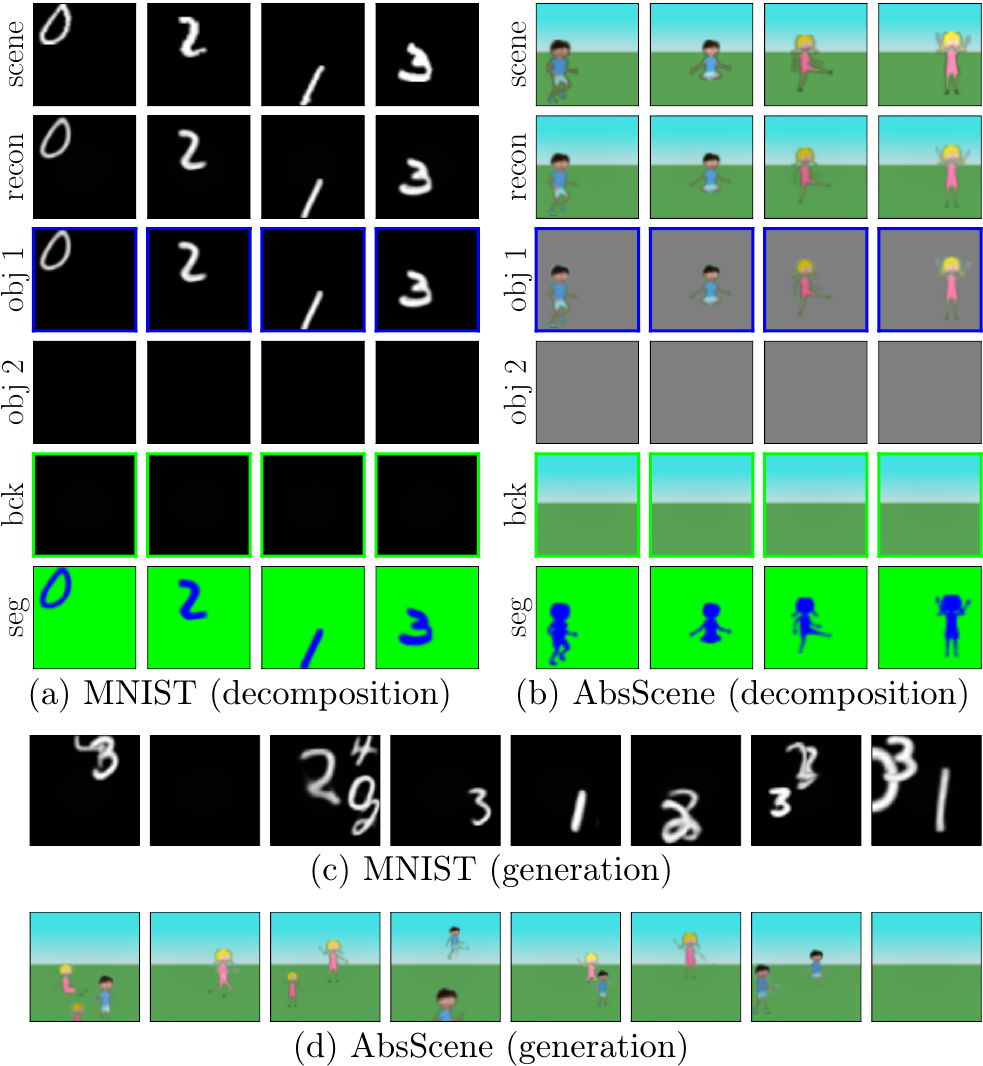}
	\caption{The decomposition and generation results of the models trained on single-object images with supervision. Reconstructed objects are superimposed on black (MNIST) or gray (AbsScene) images.}
	\label{fig:single}
\end{figure}

\begin{figure*}[!t]
	\centering
	\includegraphics[width=0.99\linewidth]{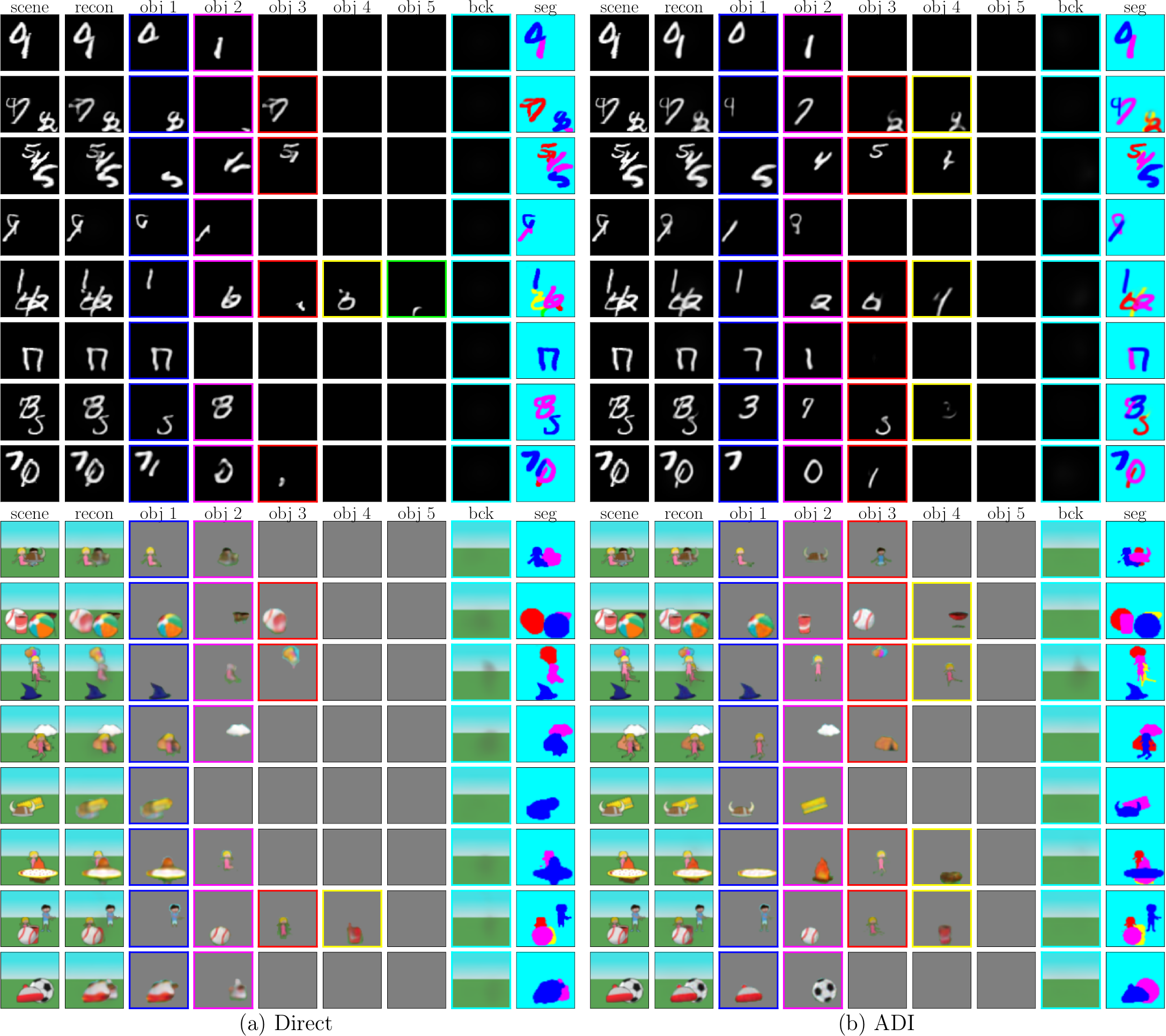}
	\caption{Decomposition results of the models trained with (ADI) or without (Direct) the acquired impressions. The average degrees of occlusion are $50\%\!\sim\!100\%$. Reconstructed objects are superimposed on black (MNIST) or gray (AbsScene) images.}
	\label{fig:compare_test}
\end{figure*}

\textbf{Datasets}:
The effectiveness of the proposed ADI framework is evaluated on two types of datasets. In the first type of datasets, images are composed of $\num[group-separator={,}]{70000}$ variants of handwritten digits $0 \!\sim\! 9$ in the MNIST dataset \cite{lecun1998gradient}. In the second type of datasets, images are composed of $70$ variants of boys and girls as well as $56$ other types of abstract objects provided by the Abstract Scene Dataset \cite{zitnick2013bringing,zitnick2013learning}. These two types of datasets are referred to as \emph{MNIST} and \emph{AbsScene}, respectively. In both types of datasets, the sizes of images are $64 \times 64$. The datasets $\mathcal{D}_{\text{single}}$ used in the first learning stage consist of $\num[group-separator={,}]{10000}$ images containing a single object. Only some of the objects may appear in $\mathcal{D}_{\text{single}}$ ($\num[group-separator={,}]{35735}$ variants of digits $0\!\sim\!4$ for the MNIST dataset, and $70$ variants of boys and girls for the AbsScene dataset). The datasets $\mathcal{D}_{\text{multi}}$ used in the second learning stage consist of $\num[group-separator={,}]{50000}$ images containing $2\!\sim\!4$ objects, and all the objects may appear in $\mathcal{D}_{\text{multi}}$. To investigate the influence of object occlusion to the effectiveness of the proposed ADI framework, each dataset is divided into $2$ subsets, which differ from each other in the average degree of occlusion ($0\%\!\sim\!50\%$ and $50\%\!\sim\!100\%$). The average degree of occlusion of each image is measured by first computing the ratio of overlapped area to the total area of bounding box for each object, and then averaging the ratios over all the objects in the image. $\num[group-separator={,}]{10000}$ images containing $2\!\sim\!4$ or $5\!\sim\!6$ objects are used to evaluate the scene decomposition performance and generalizability of the trained models. Some examples of the two types of datasets can be found in the rows and columns labeled with ``scene'' in Figures \ref{fig:single} and \ref{fig:compare_test}.

\textbf{Evaluation Metrics}:
Four metrics are used to evaluate the performance of the trained models: 1) \emph{Object Reconstruction Error} (MSE) measures the similarity between the reconstructions of discovered objects and the ground truth by mean squared error (MSE). It is evaluated on pixels belonging to the union of estimated and ground truth shapes of objects, and provides information about how accurately the occluded regions of objects are estimated. 2) \emph{Adjusted Mutual Information} (AMI) assesses qualities of segmentations. It is not evaluated in the background regions to better illustrate how accurately different objects are separated. For the MNIST dataset, overlapped regions are also excluded because all objects share similar appearances. 3) \emph{Object Counting Accuracy} (OCA) is the ratio of images in which the numbers of objects are correctly estimated. 4) \emph{Object Ordering Accuracy} (OOA) is the weighted average of accuracies of the estimated pairwise object orderings. OOA scores are not reported for the MNIST datasets because the orderings of digits are indistinguishable. Formal descriptions of evaluation metrics are provided in the supplementary material. All the models are trained once and evaluated for $5$ runs.

\begin{table}[t]
	\small
	\centering
	\addtolength{\tabcolsep}{-0pt}
	\begin{tabular}{c|C{1.0in}C{1.0in}}
		\toprule
		Dataset  &        MSE        &       OCA       \\ \midrule
		MNIST   & 5.41e-2$\pm$ 4e-4 & 0.993$\pm$ 1e-3 \\
		AbsScene & 4.76e-3$\pm$ 1e-4 & 0.992$\pm$ 8e-4 \\ \bottomrule
	\end{tabular}
	\caption{MSE and OCA scores of the models trained in the first learning stage.}
	\label{tab:performance_single}
\end{table}

\begin{table*}[t]
	\small
	\centering
	\addtolength{\tabcolsep}{-0pt}
	\begin{tabular}{c|c|c|C{1.0in}C{1.0in}C{1.0in}C{1.0in}}
		\toprule
		Dataset          &               Avg Occl               & Config &            MSE             &           AMI            &           OCA            &           OOA            \\ \midrule
		\multirow{4.5}{*}{MNIST}   &  \multirow{2}{*}{$0\%\!\sim\!50\%$}  & Direct &     2.19e-1$\pm$ 6e-4      &     0.770$\pm$ 4e-4      & \textbf{0.738}$\pm$ 1e-3 &           N/A            \\
		&                                      &  ADI   & \textbf{1.02e-1}$\pm$ 3e-4 & \textbf{0.833}$\pm$ 3e-4 &     0.704$\pm$ 2e-3      &           N/A            \\
		\cmidrule{2-7}       & \multirow{2}{*}{$50\%\!\sim\!100\%$} & Direct &     3.15e-1$\pm$ 4e-4      &     0.406$\pm$ 1e-4      &     0.399$\pm$ 2e-3      &           N/A            \\
		&                                      &  ADI   & \textbf{1.94e-1}$\pm$ 5e-4 & \textbf{0.514}$\pm$ 5e-4 & \textbf{0.579}$\pm$ 4e-3 &           N/A            \\ \midrule
		\multirow{4.5}{*}{AbsScene} &  \multirow{2}{*}{$0\%\!\sim\!50\%$}  & Direct &     2.74e-2$\pm$ 2e-4      &     0.828$\pm$ 4e-4      & \textbf{0.873}$\pm$ 1e-3 &     0.876$\pm$ 1e-3      \\
		&                                      &  ADI   & \textbf{2.04e-2}$\pm$ 1e-4 & \textbf{0.874}$\pm$ 4e-4 &     0.845$\pm$ 1e-3      & \textbf{0.888}$\pm$ 2e-3 \\
		\cmidrule{2-7}       & \multirow{2}{*}{$50\%\!\sim\!100\%$} & Direct &     5.42e-2$\pm$ 7e-5      &     0.487$\pm$ 6e-4      &     0.213$\pm$ 2e-3      &     0.748$\pm$ 1e-3      \\
		&                                      &  ADI   & \textbf{1.94e-2}$\pm$ 1e-4 & \textbf{0.752}$\pm$ 5e-4 & \textbf{0.709}$\pm$ 2e-3 & \textbf{0.911}$\pm$ 9e-4 \\ \bottomrule
	\end{tabular}
	\caption{Performance evaluated on images containing $2 \!\sim\! 4$ objects. The models are trained on images containing $2 \!\sim\! 4$ objects.}
	\label{tab:performance_test}
\end{table*}

\begin{table*}[t]
	\small
	\centering
	\addtolength{\tabcolsep}{-0pt}
	\begin{tabular}{c|c|c|C{1.0in}C{1.0in}C{1.0in}C{1.0in}}
		\toprule
		Dataset          &               Avg Occl               & Config &            MSE             &           AMI            &           OCA            &           OOA            \\ \midrule
		\multirow{4.5}{*}{MNIST}   &  \multirow{2}{*}{$0\%\!\sim\!50\%$}  & Direct &     2.37e-1$\pm$ 2e-4      &     0.719$\pm$ 4e-4      &     0.303$\pm$ 3e-3      &           N/A            \\
		&                                      &  ADI   & \textbf{1.29e-1}$\pm$ 1e-4 & \textbf{0.792}$\pm$ 2e-4 & \textbf{0.562}$\pm$ 4e-3 &           N/A            \\
		\cmidrule{2-7}       & \multirow{2}{*}{$50\%\!\sim\!100\%$} & Direct &     3.29e-1$\pm$ 3e-4      &     0.541$\pm$ 3e-4      &     0.080$\pm$ 2e-3      &           N/A            \\
		&                                      &  ADI   & \textbf{2.13e-1}$\pm$ 2e-4 & \textbf{0.630}$\pm$ 3e-4 & \textbf{0.423}$\pm$ 4e-3 &           N/A            \\ \midrule
		\multirow{4.5}{*}{AbsScene} &  \multirow{2}{*}{$0\%\!\sim\!50\%$}  & Direct &     2.29e-2$\pm$ 2e-4      &     0.832$\pm$ 4e-4      &     0.610$\pm$ 4e-3      &     0.841$\pm$ 9e-4      \\
		&                                      &  ADI   & \textbf{1.60e-2}$\pm$ 7e-5 & \textbf{0.867}$\pm$ 3e-4 & \textbf{0.661}$\pm$ 3e-3 & \textbf{0.864}$\pm$ 2e-3 \\
		\cmidrule{2-7}       & \multirow{2}{*}{$50\%\!\sim\!100\%$} & Direct &     5.90e-2$\pm$ 2e-4      &     0.562$\pm$ 3e-4      &     0.030$\pm$ 2e-3      &     0.705$\pm$ 1e-3      \\
		&                                      &  ADI   & \textbf{2.29e-2}$\pm$ 7e-5 & \textbf{0.772}$\pm$ 3e-4 & \textbf{0.450}$\pm$ 2e-3 & \textbf{0.868}$\pm$ 6e-4 \\ \bottomrule
	\end{tabular}
	\caption{Performance evaluated on images containing $5 \!\sim\! 6$ objects. The models are trained on images containing $2 \!\sim\! 4$ objects.}
	\label{tab:performance_general}
\end{table*}

\subsection{Performance of Supervised Learning}

In the first learning stage, the models are trained on images $\mathcal{D}_{\text{single}}$ containing a single object under the supervision of perceived shapes of objects and background (mixture weights of the spatial mixture models). The performance of the trained models is shown in Table \ref{tab:performance_single}. Only MSE and OCA scores are reported, because only one object exists in each image and there is no need to distinguish between or determine the orderings of different objects. Both models trained on the MNIST and AbsScene datasets are able to reconstruct the objects well (low MSE scores) and determine the number of objects in the images accurately (high OCA scores). Qualitative results of decomposed scenes in $\mathcal{D}_{\text{single}}$ and generated scenes used in the second learning stage are illustrated in Figure \ref{fig:single}. The models are able to decompose scenes into objects and background accurately, and generate images containing multiple objects with high quality.

\subsection{Effectiveness of Acquired Impressions}

In the second learning stage of the ADI framework, the models continue to learn from multi-object images $\mathcal{D}_{\text{multi}}$ with the help of the impressions acquired in the first learning stage. To verify that the acquired impressions can assist the learning of visual scenes, we also directly train models using multi-object images and make comparisons between the scene decomposition performance of the models trained with and without the acquired impressions. Quantitative results are shown in Tables \ref{tab:performance_test} and \ref{tab:performance_general}, and some of the qualitative results are demonstrated in Figure \ref{fig:compare_test}. Experimental results of influences of hyperparameters and additional qualitative results are included in the supplementary material.

The performance presented in Tables \ref{tab:performance_test} and \ref{tab:performance_general} are of the same models trained on images containing $2\!\sim\!4$ objects. In Table \ref{tab:performance_test}, the distribution of the test images is identical to the distribution of the training images. In Table \ref{tab:performance_general}, the distribution of the test images is different in that the number of objects per image is within $5\!\sim\!6$ instead of $2\!\sim\!4$. On both MNIST and AbsScene datasets, the models trained using acquired impressions (ADI) are able to reconstruct individual objects better (lower MSE scores), and distinguish between and determine the orderings of different objects more accurately (higher AMI and OOA scores) than those directly trained on multi-object images (Direct).

With the help of the previously acquired impressions, the object counting accuracies (OCA scores) are also higher when objects are heavily occluded ($50\%\!\sim\!100\%$ average degrees of occlusion). When the average degrees of occlusion are relatively small ($0\%\!\sim\!50\%$), utilizing acquired impressions increases or decreases the OCA scores on test images contain $2\!\sim\!4$ or $5\!\sim\!6$ objects. The possible reason is that the distribution of test images containing $2\!\sim\!4$ objects is identical to the training images $\mathcal{D}_{\text{multi}}$, which makes determining the numbers of objects in these test images relatively easy. The extra training images provided by the generative replay strategy are drawn from a distribution different from these test images, which improves the generalizabilities of the models on test images contain $5\!\sim\!6$ but decreases the OCA scores on these test images containing $2\!\sim\!4$ objects.

Figure \ref{fig:compare_test} presents samples of decomposition results of the models trained with or without the acquired impressions. Compared with the models directly trained on multi-objects images (subfigure (a)), the models trained using previously acquired impressions (subfigure (b)) output significantly better decomposition results. Furthermore, novel objects (digits $5 \!\sim\! 9$ and abstract objects other than boys and girls) are better discovered and separated from other objects by utilizing the previously acquired impressions, even though they are not observed in the first learning stage and supervisions of them are not available.

\section{Conclusions}

In this paper, we have proposed a human-like learning framework called Acquired Deep Impressions (ADI) to facilitate the understanding of novel visual scenes by building impressions of objects with compositional latent representations and the learned decoding and inference neural networks. The proposed ADI framework is complementary to existing methods proposed for compositional scene representation in that it provides a mechanism to effectively acquire and utilize knowledge. An existing compositional deep generative model is adapted to fall within the ADI framework, and extensive experiments are conducted using this model. We have demonstrated that the model achieves significantly better decomposition performance under the guidance of previously acquired impressions (prior knowledge) in most experimental configurations, which has validated our motivation. Incorporating more expressive impressions in the ADI framework by using structured prior distributions of latent representations could be investigated in the future.

\section{Acknowledgments}

This research was supported in part by STCSM Projects (20511100400, 18511103104), Shanghai Municipal Science and Technology Major Projects (2017SHZDZX01, 2018SHZDZX01), Shanghai Research and Innovation Functional Program (17DZ2260900), and the Program for Professor of Special Appointment (Eastern Scholar) at Shanghai Institutions of Higher Learning.

\bibliographystyle{aaai}
\bibliography{merge}

\clearpage
\appendix

\section{Formal Descriptions of Evaluation Metrics}

\subsection{Object Reconstruction Error (MSE)}

$\boldsymbol{a}_k^{i}$ and $\boldsymbol{s}_k^{i}$ are the estimated appearances and shapes of background ($k \!=\! 0$) and objects ($k \!\geq\! 1$) of the $i$th image. $\tilde{\boldsymbol{a}}_k^{i}$ and $\tilde{\boldsymbol{s}}_k^{i}$ are the ground truth appearances and shapes. $\odot$ denotes element-wise multiplication. The reconstruction $\boldsymbol{o}_k^{i}$ and ground truth $\tilde{\boldsymbol{o}}_k^{i}$ of the $k$th object ($k \!\geq\! 1$) superimposed on the background are
\begin{align}
\boldsymbol{o}_k^{i} & = \boldsymbol{s}_k^{i} \odot \boldsymbol{a}_k^{i} + (1 - \boldsymbol{s}_k^{i}) \odot \boldsymbol{a}_k^{i} \\
\tilde{\boldsymbol{o}}_k^{i} & = \tilde{\boldsymbol{s}}_k^{i} \odot \tilde{\boldsymbol{a}}_k^{i} + (1 - \tilde{\boldsymbol{s}}_k^{i}) \odot \tilde{\boldsymbol{a}}_k^{i}
\end{align}
For each object $k$, the object reconstruction error is evaluated on pixels belonging
to the union of estimated shapes $\boldsymbol{s}_k^{i}$ and ground truth shapes $\tilde{\boldsymbol{s}}_k^{i}$. The weighted average of squared differences $(\boldsymbol{o}_{k,n}^{i} - \tilde{\boldsymbol{o}}_{k,n}^{i})^2$ across all the images $i$, objects $k$, and pixels $n$ is used as the object reconstruction error:
\begin{equation}
\text{MSE} = \frac{\sum_{i,k,n}{w_{k,n}^{i} (\boldsymbol{o}_{k,n}^{i} - \tilde{\boldsymbol{o}}_{k,n}^{i})^2}}{\sum_{i,k,n}{w_{k,n}^{i}}}
\end{equation}
The weights $w_{k,n}^{i}$ in the above equation are computed by
\begin{equation}
w_{k,n}^{i} = 1 - (1 - s_{k,n}^{i}) (1 - \tilde{s}_{k,n}^{i})
\end{equation}

\subsection{Adjusted Mutual Information (AMI)}
$\boldsymbol{m}_k^{i}$ and $\tilde{\boldsymbol{m}}_k^{i}$ are the estimated and ground truth perceived shapes of layers (mixture weights of spatial mixture models) of the $i$th image. The estimated and ground truth indices of perceived layers at each pixel are computed in a greedy manner: $l_{n}^{i} = \argmax_k{m_{k,n}^{i}}$ and $\tilde{l}_{n}^{i} = \argmax_k{\tilde{m}_{k,n}^{i}}$. Adjusted Mutual Information is only evaluated on pixels belonging to foreground objects, i.e., $\Omega_{\text{obj}}^{i} = \{n: \tilde{l}_n^{i} \neq 0\}$. Let $U^{i} = \{l_n^{i}: n \in \Omega_{\text{obj}}^{i}\}$ and $V^{i} = \{\tilde{l}_n^{i}: n \in \Omega_{\text{obj}}^{i}\}$ be the collections of estimated and ground truth indices of perceived layers at foreground pixels. $I$ is the number of test images. The AMI score is computed by
\begin{equation}
\text{AMI} \!=\! \frac{1}{I} \sum_{i}\!{\frac{MI(U^{i}, V^{i}) \!-\! \mathbb{E}[MI(U^{i}, V^{i})]}{\max\{H(U^{i}), H(V^{i})\} \!-\! \mathbb{E}[MI(U^{i}, V^{i})]}}
\end{equation}
In the above equation, $MI(U^{i}, V^{i})$ is the mutual information between $U^{i}$ and $V^{i}$, and $H(U^{i})$ and $H(V^{i})$ are the entropies of $U^{i}$ and $V^{i}$.

\subsection{Object Counting Accuracy (OCA)}

$K^{i}$ and $\tilde{K}^{i}$ denote the estimated and ground truth number of objects in the $i$th test image. $I$ is the number of test images. The OCA score is computed by
\begin{equation}
\text{OCA} = \frac{1}{I} \sum_{i}{[K^{i}=\tilde{K}^{i}]}
\end{equation}
The $[\cdot]$ in the above equation is the Iverson bracket, i.e., $[K^{i}=\tilde{K}^{i}]$ is $1$ if $K^{i}$ equals $\tilde{K}^{i}$, and is $0$ otherwise.

\subsection{Object Ordering Accuracy (OOA)}

For the $i$th test image, $\boldsymbol{a}_k^{i}$ and $\boldsymbol{s}_k^{i}$ represent the estimated appearances and shapes, and $\tilde{\boldsymbol{a}}_k^{i}$ and $\tilde{\boldsymbol{s}}_k^{i}$ represent the ground truth appearances and shapes. Let $r_{k_1,k_2}^{i}$ ($k_1 < k_2$) be the variable indicating whether the relative ordering of the $k_1$th object and $k_2$th in the $i$th test image is correctly estimated, i.e., $r_{k_1,k_2}^{i}$ is $1$ if the estimated ordering is correct, and is $0$ otherwise. The weighted average of $r_{k_1,k_2}^{i}$ is used as the OOA score:
\begin{equation}
\text{OOA} = \frac{\sum_{i,k_1 < k_2}{v_{k_1,k_2}^{i} r_{k_1,k_2}^{i}}}{\sum_{i,k_1 < k_2}{v_{k_1,k_2}^{i}}}
\end{equation}
Let $n$ be the index of pixel. The weights $v_{k_1,k_2}^{i}$ in the above equation are computed by
\begin{equation}
v_{k_1,k_2}^{i} = \sum_{n}{s_{k_1,n}^{i} \, \tilde{s}_{k_2,n}^{i} \, (\boldsymbol{a}_{k_1,n}^{i} - \tilde{\boldsymbol{a}}_{k_2,n}^i)^2}
\end{equation}

\section{Adaptation of GMIOO}

\subsection{Generative Model}

The generative model of the adapted GMIOO is almost identical to the original version. There are two main differences. The first one lies in the latent variables of objects. For simplicity, the adapted GMIOO only divides the latent variable of each object into an intrinsic part (which characterizes appearance and shape) and an extrinsic part (which characterizes scale and translation), while the original version further divides the intrinsic part and represents appearance and shape separately. The second one is that the adapted GMIOO samples indicator variables $\boldsymbol{l}$ directly based on perceived shapes, instead of first sampling binary variables based on complete shapes and then transforming the binary variables into $\boldsymbol{l}$. These two differences in the generative model also result in differences in the variational inference.

In the generative model, the number of latent variables of objects, which is denoted by $K$, is sampled according to the following procedure:
\begin{gather*}
\begin{aligned}
\nu_k & \sim \Beta\big(\alpha, 1\big), && k \geq 1 \\
y_k & \sim \Bernoulli\big(\textstyle{\prod}_{k'=1}^{k}{\nu_{k'}}\big), && k \geq 1
\end{aligned} \\
K = \sum\nolimits_{k=1}^{\infty}{y_k}
\end{gather*}

Both latent variables of background $\boldsymbol{z}_0$ and the intrinsic part of latent variables of objects $\boldsymbol{z}_k^{\text{int}}$ ($1 \leq k \leq K$) are assumed to be drawn from standard normal distributions. The extrinsic part of latent variables of objects $\boldsymbol{z}_k^{\text{ext}}$ ($1 \leq k \leq K$) is assumed to be drawn from a normal distribution with tunable hyperparameters. Detailed expressions of prior distributions of background $p(\boldsymbol{z}_k; \boldsymbol{\theta}_{\text{bck}})$ and objects $p(\boldsymbol{z}_k; \boldsymbol{\theta}_{\text{obj}})$ are
\begin{gather*}
p(\boldsymbol{z}_k; \boldsymbol{\theta}_{\text{bck}}) = \mathcal{N}\big(\boldsymbol{z}_k; \boldsymbol{0}, \boldsymbol{I}\big) \\
p(\boldsymbol{z}_k; \boldsymbol{\theta}_{\text{obj}}) = \mathcal{N}\big(\boldsymbol{z}_k^{\text{int}}; \boldsymbol{0}, \boldsymbol{I}\big) \, \mathcal{N}\big(\boldsymbol{z}_k^{\text{ext}}; \boldsymbol{\mu}^{\text{ext}}, \diag(\boldsymbol{\sigma}^{\text{ext}})^2\big)
\end{gather*}

The function $f_{\text{bck}}$ which transforms the latent variable of background into appearance and complete shape consists of a neural network $f_{\text{apc}}^{\text{bck}}$ and a manually defined mapping which constantly outputs $\boldsymbol{1}$. The function $f_{\text{obj}}$ which transforms latent variables of objects into appearances and complete shapes consists of two neural networks $f_{\text{apc}}^{\text{obj}}$ and $f_{\text{shp}}$, as well as a manually defined mapping $f_{\text{stn}}$ which applies scale and translation transformations to the outputs of $f_{\text{apc}}^{\text{obj}}$ and $f_{\text{shp}}$ based on the extrinsic part of latent variables of objects. The appearances $\boldsymbol{a}$ and complete shapes $\boldsymbol{s}$ of background ($k = 0$) and objects ($1 \!\leq\! k \!\leq\! K$) are computed by
\begin{align*}
\boldsymbol{a}_{k,n} & =
\begin{cases}
f_{\text{apc}}^{\text{bck}}(\boldsymbol{z}_k), \\
f_{\text{stn}}\big(f_{\text{apc}}^{\text{obj}}(\boldsymbol{z}_k^{\text{int}}), \boldsymbol{z}_k^{\text{ext}}\big),
\end{cases} &&
\begin{aligned}
& k = 0 \\
& 1 \leq k \leq K
\end{aligned} \\
s_{k,n} & =
\begin{cases}
1, \\
f_{\text{stn}}\big(f_{\text{shp}}(\boldsymbol{z}_k^{\text{int}}), \boldsymbol{z}_k^{\text{ext}}\big),
\end{cases} &&
\begin{aligned}
& k = 0 \\
& 1 \leq k \leq K
\end{aligned}
\end{align*}

The complete shapes of background and objects are transformed into perceived shapes $\boldsymbol{m}$ using the compositing function $f_{\text{comp}}$. The detailed expression of computation is
\begin{equation*}
m_{k,n} =
\begin{cases}
s_{k,n} \, \textstyle{\prod}_{k'=1}^{K}(1 - s_{k',n}), & k = 0 \\
s_{k,n} \, \textstyle{\prod}_{k'=1}^{k-1}(1 - s_{k',n}), & 1 \leq k \leq K
\end{cases}
\end{equation*}
The variable $l_n$ which indicates the observed layer at the $n$th pixel is then drawn from a categorical distribution parameterized by the perceived shapes
\begin{equation*}
l_n \sim \Cat(m_{0,n}, m_{1,n}, \dots, m_{K,n})
\end{equation*}
Finally, each pixel $\boldsymbol{x}_n$ is sampled from a normal distribution $p(\boldsymbol{x}_n; \boldsymbol{a}_{l_n,n})$, where
\begin{equation*}
p(\boldsymbol{x}_n; \boldsymbol{a}_{l_n,n}) = \mathcal{N}\big(\boldsymbol{x}_n; \boldsymbol{a}_{l_n,n}, \sigma_{\text{x}}^2 \boldsymbol{I}\big)
\end{equation*}
The $\sigma_{\text{x}}$ in the above expression is a tunable hyperparameter.

\subsection{Variational Inference}

Similar to the original GMIOO, the variational distribution $q(\boldsymbol{z}, \boldsymbol{\nu}, \boldsymbol{y}, \boldsymbol{l}|\boldsymbol{x})$ is factorized as
\begin{align*}
& q(\boldsymbol{z}, \boldsymbol{\nu}, \boldsymbol{y}, \boldsymbol{l}|\boldsymbol{x}) = q(\boldsymbol{z}_0|\boldsymbol{x}) \prod_{k=1}^{K_{\text{max}}}{q(\boldsymbol{z}_k^{\text{ext}}|\boldsymbol{x}) q(\boldsymbol{z}_k^{\text{int}}|\boldsymbol{z}_k^{\text{ext}}, \boldsymbol{x})} \\
& \qquad \prod_{k=1}^{K_{\text{max}}}{q(\nu_k|\boldsymbol{z}_k^{\text{ext}}, \boldsymbol{x}) q(y_k|\boldsymbol{z}_k^{\text{ext}}, \boldsymbol{x})} \prod_{n=1}^{N}{q(l_n|\boldsymbol{z}, \boldsymbol{y}, \boldsymbol{x})}
\end{align*}
In the above equation, $K_{\text{max}}$ is a hyperparameter which defines the upper bound of the number of objects that may be discovered in the image, and can be set based on the affordable computational resources. Detailed forms of the distributions on the right-hand side of the equation are
\begin{align*}
q(\boldsymbol{z}_0|\boldsymbol{x}) & = \mathcal{N}\big(\boldsymbol{z}_0; \hat{\boldsymbol{\mu}}_0, \diag(\hat{\boldsymbol{\sigma}}_0)^2\big) \\
q(\boldsymbol{z}_k^{\text{ext}}|\boldsymbol{x}) & = \mathcal{N}\big(\boldsymbol{z}_k^{\text{ext}}; \hat{\boldsymbol{\mu}}_k^{\text{ext}}, \diag(\hat{\boldsymbol{\sigma}}_k^{\text{ext}})^2\big) \\
q(\boldsymbol{z}_k^{\text{int}}|\boldsymbol{z}_k^{\text{ext}}, \boldsymbol{x}) & = \mathcal{N}\big(\boldsymbol{z}_k^{\text{int}}; \hat{\boldsymbol{\mu}}_k^{\text{int}}, \diag(\hat{\boldsymbol{\sigma}}_k^{\text{int}})^2\big) \\
q(\nu_k|\boldsymbol{z}_k^{\text{ext}}, \boldsymbol{x}) & = \Beta\big(\nu_k; \tau_{k,1}, \tau_{k,2}\big) \\
q(y_k|\boldsymbol{z}_k^{\text{ext}}, \boldsymbol{x}) & = \Bernoulli\big(y_k; \zeta_k\big) \\
q(l_n|\boldsymbol{z}, \boldsymbol{y}, \boldsymbol{x}) & = \Cat\big(l_n; \xi_{0,n}, \dots, \xi_{K_{\text{max}},n}\big)
\end{align*}
The parameters $\hat{\boldsymbol{\mu}}_0$, $\hat{\boldsymbol{\sigma}}_0$, $\hat{\boldsymbol{\mu}}_k^{\text{ext}}$, $\hat{\boldsymbol{\sigma}}_k^{\text{ext}}$, $\hat{\boldsymbol{\mu}}_k^{\text{int}}$, $\hat{\boldsymbol{\sigma}}_k^{\text{int}}$, $\tau_{k,1}$, $\tau_{k,2}$, $\zeta_k$ ($1 \!\leq\! k \!\leq\! K_{\text{max}}$) are all estimated by inference networks. For simplicity, the parameters $\xi_{k,n}$ ($0 \!\leq\! k \!\leq\! K_{\text{max}}, 1 \!\leq\! n \!\leq\! N$) are not estimated by an extra inference network, but computed according to the following expressions:
\begin{gather*}
\pi_{k,n} =
\begin{cases}
\textstyle{\prod}_{k'=1}^{K}(1 - y_{k'} s_{k',n}), & k = 0 \\
y_k s_{k,n} \, \textstyle{\prod}_{k'=1}^{k-1}(1 - y_{k'} s_{k',n}), & 1 \leq k \leq K_{\text{max}}
\end{cases} \\
\xi_{k,n} = \frac{\pi_{k,n} \cdot \mathcal{N}\big(\boldsymbol{x}_n; \boldsymbol{a}_{k,n}, \sigma_{\text{x}}^2 \boldsymbol{I}\big)}{\sum_{k'=0}^{K_{\text{max}}}{\pi_{k',n} \cdot \mathcal{N}\big(\boldsymbol{x}_n; \boldsymbol{a}_{k',n}, \sigma_{\text{x}}^2 \boldsymbol{I}\big)}}
\end{gather*}
The appearances $\boldsymbol{a}_k$ and the complete shapes $\boldsymbol{s}_k$ are computed following the generative model. The latent variables $y_k$ are sampled from $q(y_k|\boldsymbol{z}_k^{\text{ext}}, \boldsymbol{x})$. The estimated number of objects is computed by $\hat{K} = \sum_{k=1}^{K_{\text{max}}}{y_k}$. The prior of the number of objects favors small values, and the KL divergence between the variational and prior distributions discourages the same object to be discovered repeatedly.

Same as the original GMIOO, parameters of the variational distribution are initialized sequentially by initialization neural networks. Different from the original version, these parameters are updated in parallel after the initialization, in order to speed up the inference of latent variables and the training of neural networks. The ordering of latent variables of objects is also concurrently updated with parameters of the variational distribution, because the inference results may get stuck in undesired local optimum if the ordering determined by the initialization neural networks is kept unmodified during the inference. The possible number of ordering is $K_{\text{max}}!$, which makes it infeasible to enumerate the ordering if $K_{\text{max}}$ is large. Therefore, the ordering of latent variables of objects is updated in a heuristic and greedy manner, based on the similarities between the image $\boldsymbol{x}$ and appearances of objects $\boldsymbol{a}_k$ in the regions masked by complete shapes of objects $\boldsymbol{s}_k$. The time complexity of updating the ordering in this way is proportional to $K_{\text{max}}$.

\section{Choices of Hyperparameters}

\subsection{Base Model (GMIOO)}

The likelihood function is a normal distribution with standard deviation $0.3$. The priors of latent variables of relative scales and translations of objects are normal distributions with mean $0$ and standard deviation $0.5$. The priors of latent variables of background as well as shapes and appearances are standard normal distributions. The $\alpha$ parameter of the beta distribution is $3.0$. Each latent representation is updated for $2$ iterations. The batch size is chosen to be $64$, and the learning rate is $2 \!\times\! 10^{-4}$. When training the models, the upper bound of the number of objects $K_{\text{max}}$ is chosen to be $2$ and $5$ during the first and the second learning stages, respectively. When testing the models on images containing $2 \!\sim\! 4$ and $5 \!\sim\! 6$ objects, $K_{\text{max}}$ is chosen to be $5$ and $7$, respectively. Detailed choices of other hyperparameters such as the hyperparameters of neural networks are included in the provided source code.

\subsection{Learning Framework (ADI)}

In the first learning stage of ADI, the models are trained on images containing a single object for $500$ epochs. $\beta_{\text{stg1}}$ is increased linearly from $0$ (in the $1$st epoch) to $10.0$ (in the $500$th epoch) as the epoch increases. $\alpha_{\text{stg1}}$ is chosen to be $1.0$. In the second learning stage of ADI, the models continue to learn on images containing multiple objects for $100$ epochs. $\beta_{\text{stg2}}$ is chosen to be $1.0$, and $\alpha_{\text{stg2}}$ is set to $1.0$. For the models directly trained on multi-object images, the number of epochs is also chosen to be $100$.

\section{Influences of Hyperparameters}

To investigate the influences of hyperparameters on the performance of the models, different values of $\beta_{\text{stg1}}$, $\alpha_{\text{stg1}}$, $\beta_{\text{stg2}}$, and $\alpha_{\text{stg2}}$ are chosen.

\begin{figure*}[t]
	\centering
	\includegraphics[width=0.99\linewidth]{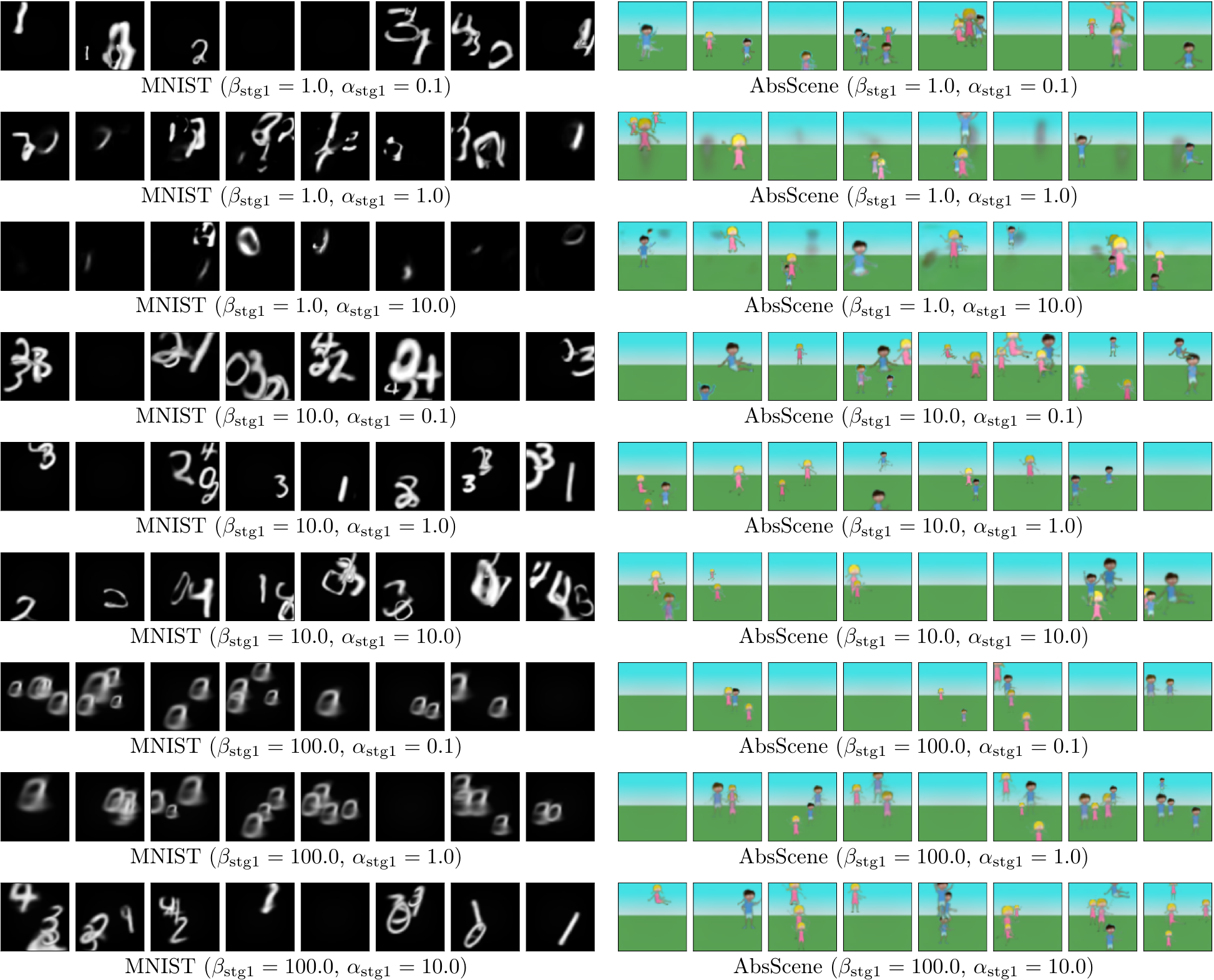}
	\caption{Scenes generated by the models trained using different hyperparameters in the first learning stage. The base model is GMIOO.}
	\label{fig:gmioo_compare_gen}
\end{figure*}

In the first learning stage, the models are trained using $9$ combinations of hyperparameters $\beta_{\text{stg1}} \!\in\! \{1.0, 10.0, 100.0\}$ and $\alpha_{\text{stg1}} \!\in\! \{0.1, 1.0, 10.0\}$. During the training, $\beta_{\text{stg1}}$ is increased linearly from $0$ to the target value, and $\alpha_{\text{stg1}}$ is kept constant. The quantitative results are shown in Table \ref{tab:gmioo_compare_single}, and the qualitative results are shown in Figure \ref{fig:gmioo_compare_gen}. In general, the reconstruction error (MSE) becomes larger as $\beta_{\text{stg1}}$ increases, and the object counting accuracy (OCA) is insensitive to the choices of $\beta_{\text{stg1}}$ and $\alpha_{\text{stg1}}$. The hyperparameters used in the first learning stage are not chosen solely based on the MSE and OCA scores. The qualities of the generated samples, which can be assessed by visual inspection, are also important to the choices of hyperparameters.

In the second learning stage, the models continue to learn from multi-object images based on the models trained with $\beta_{\text{stg1}} \!=\! 10.0$ and $\alpha_{\text{stg1}} \!=\! 1.0$ in the first learning stage, using $8$ combinations of hyperparameters $\beta_{\text{stg2}} \!\in\! \{1.0, 10.0\}$ and $\alpha_{\text{stg2}} \!\in\! \{0.0, 0.1, 1.0, 10.0\}$. The quantitative results are shown in Table \ref{tab:gmioo_compare_test} (tested on images containing $2 \!\sim\! 4$ objects) and Table \ref{tab:gmioo_compare_general} (tested on images containing $5 \!\sim\! 6$ objects).

\begin{table}[t]
	\small
	\centering
	\addtolength{\tabcolsep}{-1.5pt}
	\begin{tabular}{c|c|c|C{0.8in}C{0.8in}}
		\toprule
		Dataset          & $\beta_{\text{stg1}}$ & $\alpha_{\text{stg1}}$ &        MSE        &       OCA       \\ \midrule
		\multirow{9}{*}{MNIST}   &          1.0          &          0.1           & 2.67e-2$\pm$ 5e-4 & 0.994$\pm$ 4e-4 \\
		&          1.0          &          1.0           & 2.41e-2$\pm$ 3e-4 & 0.993$\pm$ 5e-4 \\
		&          1.0          &          10.0          & 2.09e-2$\pm$ 2e-4 & 0.995$\pm$ 8e-4 \\
		&         10.0          &          0.1           & 6.44e-2$\pm$ 5e-4 & 0.993$\pm$ 7e-4 \\
		&         10.0          &          1.0           & 5.41e-2$\pm$ 4e-4 & 0.993$\pm$ 1e-3 \\
		&         10.0          &          10.0          & 4.01e-2$\pm$ 6e-4 & 0.994$\pm$ 8e-4 \\
		&         100.0         &          0.1           & 1.88e-1$\pm$ 8e-5 & 0.992$\pm$ 4e-4 \\
		&         100.0         &          1.0           & 1.82e-1$\pm$ 3e-4 & 0.993$\pm$ 7e-4 \\
		&         100.0         &          10.0          & 8.85e-2$\pm$ 2e-4 & 0.994$\pm$ 3e-4 \\ \midrule
		\multirow{9}{*}{AbsScene} &          1.0          &          0.1           & 2.50e-3$\pm$ 1e-4 & 0.992$\pm$ 1e-3 \\
		&          1.0          &          1.0           & 3.01e-3$\pm$ 6e-5 & 0.992$\pm$ 7e-4 \\
		&          1.0          &          10.0          & 2.88e-3$\pm$ 2e-5 & 0.997$\pm$ 3e-4 \\
		&         10.0          &          0.1           & 3.57e-3$\pm$ 3e-5 & 0.991$\pm$ 5e-4 \\
		&         10.0          &          1.0           & 4.76e-3$\pm$ 1e-4 & 0.992$\pm$ 8e-4 \\
		&         10.0          &          10.0          & 4.82e-3$\pm$ 6e-5 & 0.997$\pm$ 5e-4 \\
		&         100.0         &          0.1           & 1.60e-2$\pm$ 5e-5 & 0.992$\pm$ 1e-3 \\
		&         100.0         &          1.0           & 1.39e-2$\pm$ 1e-4 & 0.993$\pm$ 6e-4 \\
		&         100.0         &          10.0          & 6.86e-3$\pm$ 6e-5 & 0.997$\pm$ 4e-4 \\ \bottomrule
	\end{tabular}
	\caption{MSE and OCA scores of the models trained using different hyperparameters in the first learning stage. The base model is GMIOO.}
	\label{tab:gmioo_compare_single}
\end{table}

\section{Running Time Complexity}

Let $T$ denote the number of iterative updates of latent variables, and $K_{\text{max}}$ denote the upper bound of the number of objects during the variational inference. $C$ is the ratio of the number of generated images with pseudo-labels to the number of images provided by the dataset in each mini-batch. $S$ is the number of training epochs. When training the models, the time complexities of the original GMIOO and the adapted version are $O(S \cdot (T + 1) \cdot (K_{\text{max}} + 1))$ and $O(S \cdot (C + 1) \cdot (T + 1) \cdot (K_{\text{max}} + 1))$, respectively. When testing the models, the time complexities of the original and adapted versions are both $O((T + 1) \cdot (K_{\text{max}} + 1))$.

\section{Computing Infrastructure}

All the models are trained on a single NVIDIA GeForce RTX 2080 Ti GPU. The CPU used is Intel Xeon CPU E5-2678 v3, and the size of memory is 256G. The operating system is Ubuntu 16.04. The code is written using the PyTorch \cite{paszke2019pytorch} deep learning framework, and the version of PyTorch is 1.6.

\section{Additional Qualitative Results}

\begin{figure*}[t]
	\centering
	\includegraphics[width=0.99\linewidth]{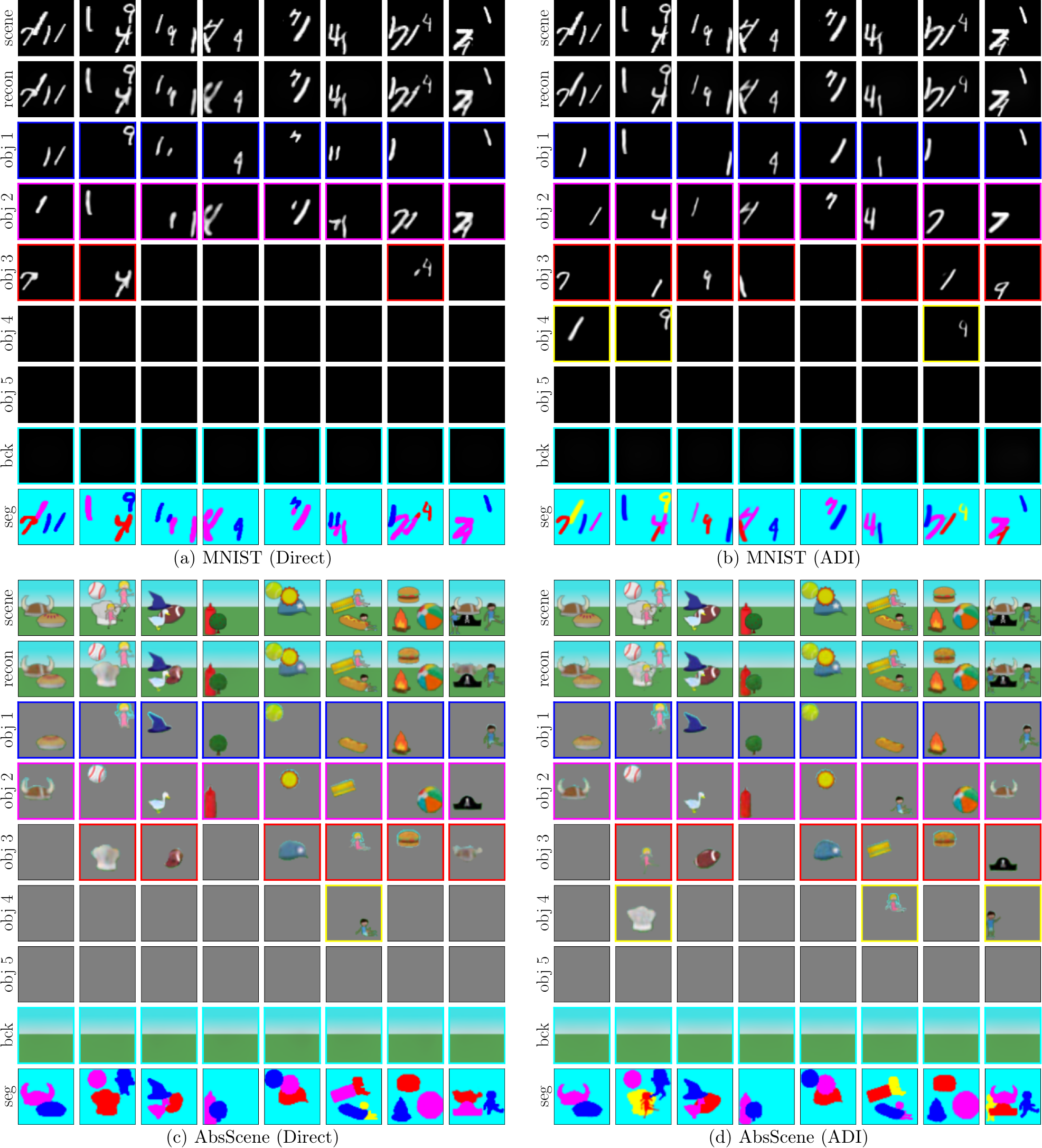}
	\caption{Decomposition results of the models trained with (ADI) or without (Direct) the acquired impressions. Each test image contains $2\!\sim\!4$ objects, and the average degrees of occlusion are $0\%\!\sim\!50\%$. Reconstructed objects are superimposed on black (MNIST) or gray (AbsScene) images.}
	\label{fig:compare_test_0}
\end{figure*}

\begin{figure*}[t]
	\centering
	\includegraphics[width=0.89\linewidth]{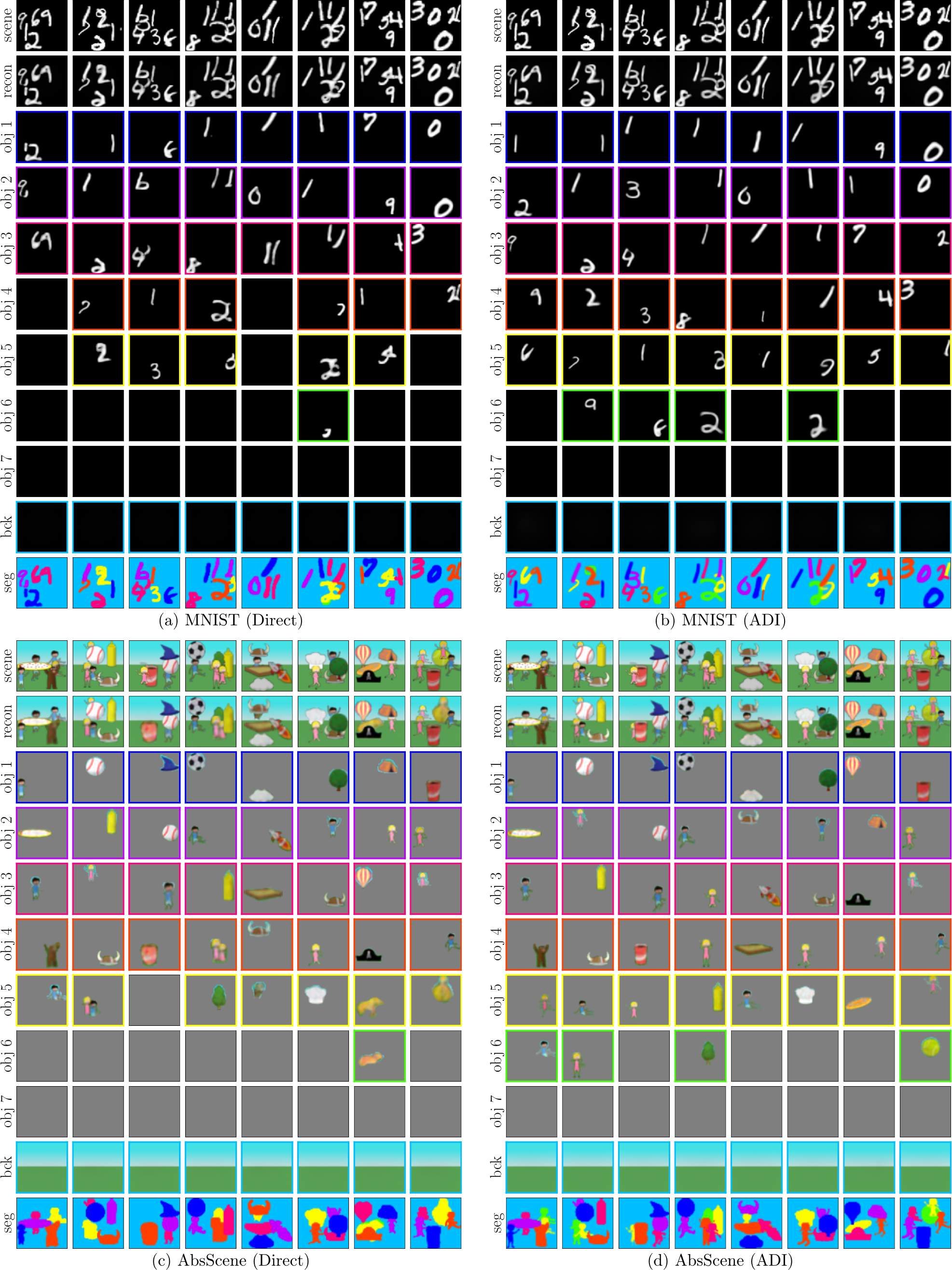}
	\caption{Decomposition results of the models trained with (ADI) or without (Direct) the acquired impressions. Each test image contains $5\!\sim\!6$ objects, and the average degrees of occlusion are $0\%\!\sim\!50\%$. Reconstructed objects are superimposed on black (MNIST) or gray (AbsScene) images.}
	\label{fig:compare_general_0}
\end{figure*}

\begin{figure*}[t]
	\centering
	\includegraphics[width=0.89\linewidth]{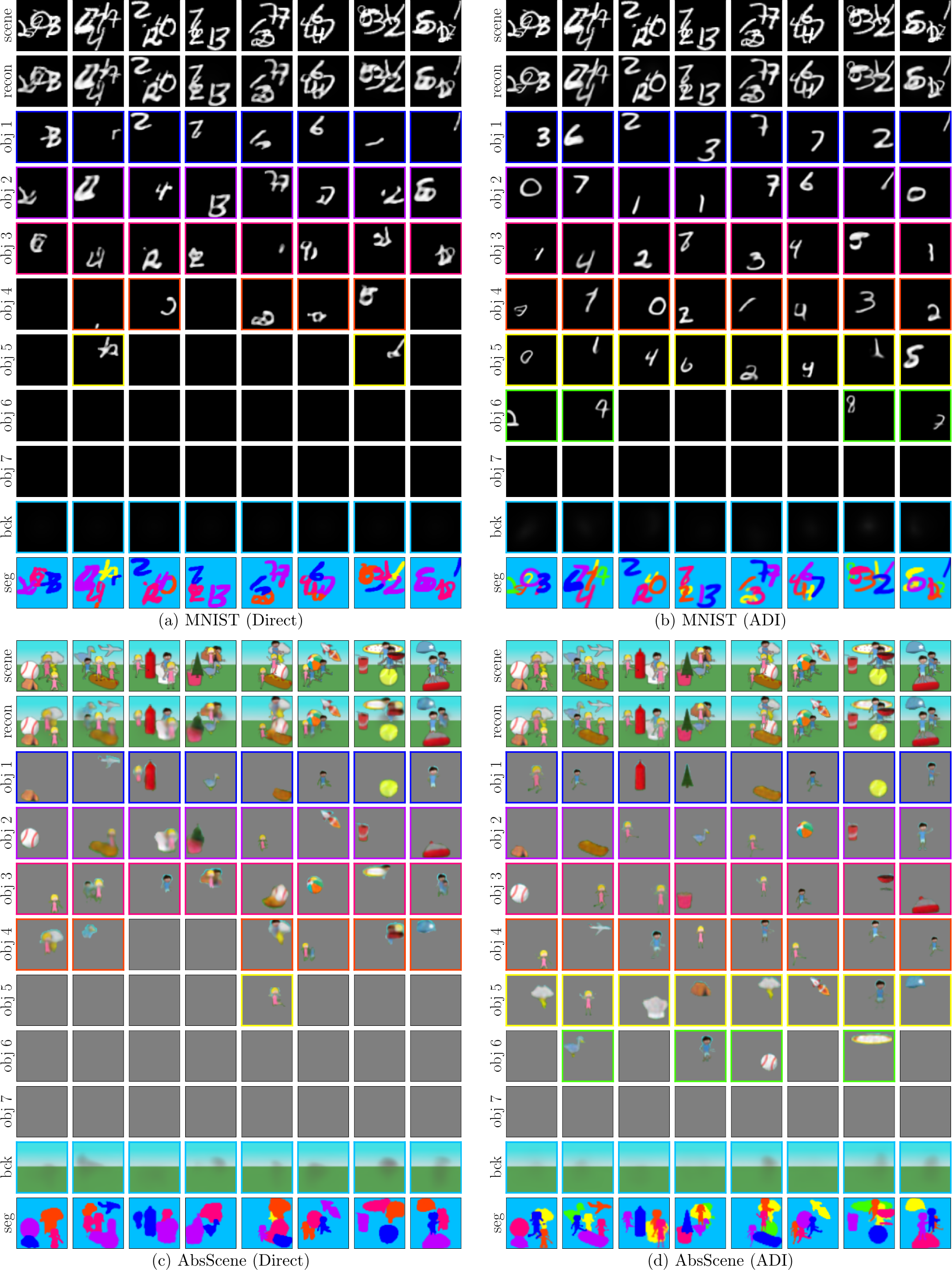}
	\caption{Decomposition results of the models trained with (ADI) or without (Direct) the acquired impressions. Each test image contains $5\!\sim\!6$ objects, and the average degrees of occlusion are $50\%\!\sim\!100\%$. Reconstructed objects are superimposed on black (MNIST) or gray (AbsScene) images.}
	\label{fig:compare_general_1}
\end{figure*}

Additional qualitative results are presented in Figures \ref{fig:compare_test_0}, \ref{fig:compare_general_0}, and \ref{fig:compare_general_1}. It is shown that utilizing previously acquired impressions improves decomposition results noticeably in most experimental configurations.

\section{Additional Experiments of Adapting AIR}

\subsection{Adaptation of AIR}

\subsubsection{Generative Model}

Three major modifications are made to the generative model of AIR to make it fall within the ADI framework. The first modification is to add an extra latent variable representing the background of image. This modification makes the model capable of generating background, so that manually providing background is not necessary. The second modification is to model images with spatial mixture models, instead of summations of layers. This modification separates the appearance and shape attributes of objects, which enables the models to produce pixel-level segmentations of images and also improves the qualities of generated images. The third modification is to replace the geometric prior on the number of objects with the prior used in GMIOO. This modification breaks the direct dependencies among the latent variables representing presences of objects, and leads to higher object counting accuracies in the second learning stage. The resultant generative model is identical to the adapted version of GMIOO.

\subsubsection{Variational Inference}

The inference of latent variables is almost identical to the adapted GMIOO. The only two differences are that the adapted AIR does not iteratively update latent variables during the inference, and uses slightly different inference networks because iterative updates are not needed. Compared to the original AIR, an extra neural network is added to infer the latent variable of background, and the structures of neural networks for inferring latent variables representing \emph{presence}, \emph{where}, and \emph{what} attributes of objects are also different.

\begin{table}[t]
	\small
	\centering
	\addtolength{\tabcolsep}{-1.5pt}
	\begin{tabular}{c|c|c|C{0.8in}C{0.8in}}
		\toprule
		Dataset          & $\beta_{\text{stg1}}$ & $\alpha_{\text{stg1}}$ &        MSE        &       OCA        \\ \midrule
		\multirow{9}{*}{MNIST}   &          1.0          &          0.1           & 2.69e-2$\pm$ 7e-5 & 1.000$\pm$ 5e-5  \\
		&          1.0          &          1.0           & 2.61e-2$\pm$ 3e-5 & 1.000$\pm$ 1e-4  \\
		&          1.0          &          10.0          & 2.28e-2$\pm$ 2e-5 & 1.000$\pm$ 1e-4  \\
		&         10.0          &          0.1           & 6.50e-2$\pm$ 1e-4 & 1.000$\pm$ 1e-4  \\
		&         10.0          &          1.0           & 5.77e-2$\pm$ 2e-4 & 1.000$\pm$ 1e-4  \\
		&         10.0          &          10.0          & 4.50e-2$\pm$ 8e-5 & 1.000$\pm$ 1e-4  \\
		&         100.0         &          0.1           & 1.90e-1$\pm$ 3e-5 & 0.999$\pm$ 2e-4  \\
		&         100.0         &          1.0           & 1.75e-1$\pm$ 2e-4 & 1.000$\pm$ 6e-5  \\
		&         100.0         &          10.0          & 8.55e-2$\pm$ 9e-5 & 1.000$\pm$ 1e-4  \\ \midrule
		\multirow{9}{*}{AbsScene} &          1.0          &          0.1           & 1.88e-3$\pm$ 5e-6 & 0.991$\pm$ 3e-4  \\
		&          1.0          &          1.0           & 2.65e-3$\pm$ 4e-6 & 1.000$\pm$ 9e-5  \\
		&          1.0          &          10.0          & 3.08e-3$\pm$ 6e-6 & 1.000$\pm$ 4e-5  \\
		&         10.0          &          0.1           & 4.65e-3$\pm$ 3e-5 & 0.991$\pm$ 4e-4  \\
		&         10.0          &          1.0           & 4.22e-3$\pm$ 1e-5 & 1.000$\pm$ 6e-5  \\
		&         10.0          &          10.0          & 4.59e-3$\pm$ 1e-5 & 1.000$\pm$ 4e-5  \\
		&         100.0         &          0.1           & 1.50e-2$\pm$ 7e-5 & 1.000$\pm$ 0e-0 \\
		&         100.0         &          1.0           & 1.49e-2$\pm$ 1e-4 & 1.000$\pm$ 8e-5  \\
		&         100.0         &          10.0          & 7.18e-3$\pm$ 5e-5 & 1.000$\pm$ 8e-5  \\ \bottomrule
	\end{tabular}
	\caption{MSE and OCA scores of the models trained using different hyperparameters in the first learning stage. The base model is AIR.}
	\label{tab:air_compare_single}
\end{table}

\subsection{Choices of Hyperparameters}

The hyperparameters of the base model (AIR) and the learning framework (ADI) are almost identical to the ones used in the adapted GMIOO. The only difference is that the adapted AIR does not use a hyperparameter to represent the number of iterative updates of latent variables.

\subsection{Effectiveness of Acquired Impressions}

The performance evaluated under different configurations is shown in Table \ref{tab:air_performance_test} (tested on images containing $2 \!\sim\! 4$ objects) and Table \ref{tab:air_performance_general} (tested on images containing $5 \!\sim\! 6$ objects). In the \emph{Basic} configuration, only the first modification in the adaptation of AIR (i.e., add an extra latent variable of background) is made, and the models are directly trained on multi-object images. Compared to the other two configurations, the \emph{Basic} configuration is more similar to the original AIR. Because the complete shapes of objects are not estimated by the models in this configuration, the MSE scores cannot be computed. Because there is no natural way to determine the perceived shapes of objects, the AMI scores are computed based on the pixel-level segmentations of images that are heuristically estimated according to the similarities between the observed image and the summation of the reconstructed background and each reconstructed object. In the \emph{Direct} configuration, all the three modifications are made, and the models are also directly trained on multi-object images. In the \emph{ADI} configuration, the fully adapted models are trained on multi-object images with the help of impressions that are acquired in the first learning stage. The models trained under the ADI configuration achieve the best performance in all the settings.

\subsection{Influences of Hyperparameters}

\begin{figure*}[t]
	\centering
	\includegraphics[width=0.99\linewidth]{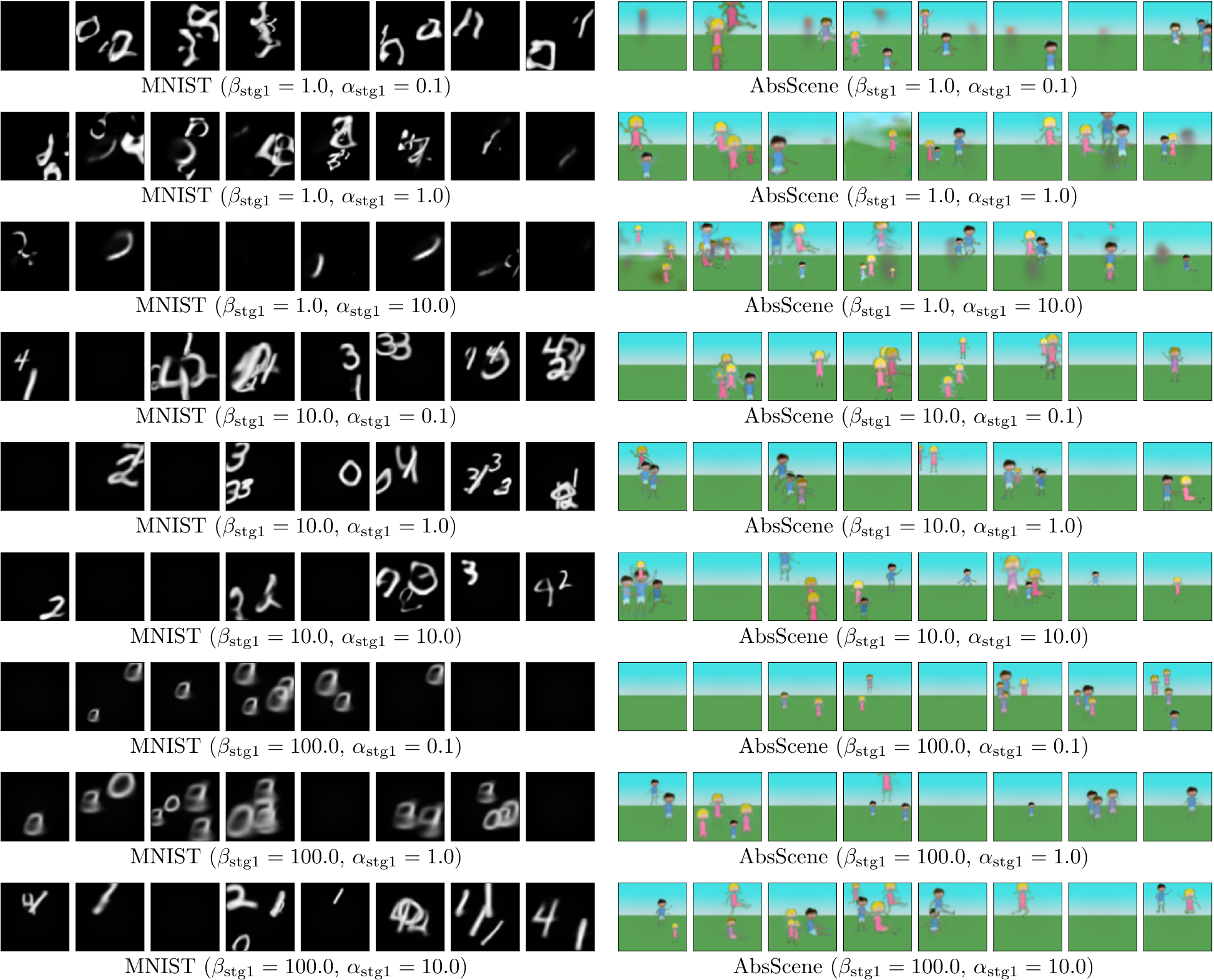}
	\caption{Scenes generated by the models trained using different hyperparameters in the first learning stage. The base model is AIR.}
	\label{fig:air_compare_gen}
\end{figure*}

Same as the adapted GMIOO, the adapted AIR is trained using different values of $\beta_{\text{stg1}}$, $\alpha_{\text{stg1}}$, $\beta_{\text{stg2}}$, and $\alpha_{\text{stg2}}$ to investigate the influences of hyperparameters. The quantitative results and qualitative results of the first learning stage are shown in Table \ref{tab:air_compare_single} and Figure \ref{fig:air_compare_gen}, respectively. The quantitative results of the second learning stage are shown in Table \ref{tab:air_compare_test} (tested on images containing $2 \!\sim\! 4$ objects) and Table \ref{tab:air_compare_general} (tested on images containing $5 \!\sim\! 6$ objects).

\clearpage

\begin{table*}[t]
	\small
	\centering
	\addtolength{\tabcolsep}{-0pt}
	\begin{tabular}{c|c|c|c|C{0.95in}C{0.95in}C{0.95in}C{0.95in}}
		\toprule
		Dataset            &               Avg Occl               & $\beta_{\text{stg2}}$ & $\alpha_{\text{stg2}}$ &       MSE        &      AMI       &      OCA       &      OOA       \\ \midrule
		\multirow{16.5}{*}{MNIST}   &  \multirow{8}{*}{$0\%\!\sim\!50\%$}  &          1.0          &          0.0           & 1.11e-1$\pm$ 3e-4 & 0.783$\pm$ 4e-4 & 0.340$\pm$ 1e-3 &      N/A       \\
		&                                      &          1.0          &          0.1           & 9.62e-2$\pm$ 2e-4 & 0.805$\pm$ 4e-4 & 0.377$\pm$ 4e-3 &      N/A       \\
		&                                      &          1.0          &          1.0           & 1.02e-1$\pm$ 3e-4 & 0.833$\pm$ 3e-4 & 0.704$\pm$ 2e-3 &      N/A       \\
		&                                      &          1.0          &          10.0          & 1.03e-1$\pm$ 4e-4 & 0.833$\pm$ 5e-4 & 0.659$\pm$ 3e-3 &      N/A       \\
		&                                      &         10.0          &          0.0           & 1.42e-1$\pm$ 3e-4 & 0.767$\pm$ 7e-4 & 0.536$\pm$ 4e-3 &      N/A       \\
		&                                      &         10.0          &          0.1           & 1.50e-1$\pm$ 3e-4 & 0.795$\pm$ 3e-4 & 0.700$\pm$ 2e-3 &      N/A       \\
		&                                      &         10.0          &          1.0           & 1.52e-1$\pm$ 4e-4 & 0.813$\pm$ 3e-4 & 0.726$\pm$ 3e-3 &      N/A       \\
		&                                      &         10.0          &          10.0          & 1.58e-1$\pm$ 5e-4 & 0.819$\pm$ 8e-4 & 0.776$\pm$ 4e-3 &      N/A       \\
		\cmidrule{2-8}        & \multirow{8}{*}{$50\%\!\sim\!100\%$} &          1.0          &          0.0           & 2.37e-1$\pm$ 2e-4 & 0.391$\pm$ 9e-4 & 0.436$\pm$ 4e-3 &      N/A       \\
		&                                      &          1.0          &          0.1           & 2.00e-1$\pm$ 3e-4 & 0.453$\pm$ 5e-4 & 0.503$\pm$ 4e-3 &      N/A       \\
		&                                      &          1.0          &          1.0           & 1.94e-1$\pm$ 5e-4 & 0.514$\pm$ 5e-4 & 0.579$\pm$ 4e-3 &      N/A       \\
		&                                      &          1.0          &          10.0          & 1.65e-1$\pm$ 3e-4 & 0.560$\pm$ 4e-4 & 0.571$\pm$ 4e-3 &      N/A       \\
		&                                      &         10.0          &          0.0           & 2.19e-1$\pm$ 8e-5 & 0.387$\pm$ 4e-4 & 0.464$\pm$ 2e-3 &      N/A       \\
		&                                      &         10.0          &          0.1           & 2.03e-1$\pm$ 3e-4 & 0.435$\pm$ 1e-3 & 0.558$\pm$ 4e-3 &      N/A       \\
		&                                      &         10.0          &          1.0           & 1.98e-1$\pm$ 3e-4 & 0.482$\pm$ 5e-4 & 0.559$\pm$ 2e-3 &      N/A       \\
		&                                      &         10.0          &          10.0          & 2.11e-1$\pm$ 4e-4 & 0.492$\pm$ 4e-4 & 0.587$\pm$ 3e-3 &      N/A       \\ \midrule
		\multirow{16.5}{*}{AbsScene} &  \multirow{8}{*}{$0\%\!\sim\!50\%$}  &          1.0          &          0.0           & 2.17e-2$\pm$ 1e-4 & 0.869$\pm$ 2e-4 & 0.878$\pm$ 1e-3 & 0.891$\pm$ 1e-3 \\
		&                                      &          1.0          &          0.1           & 2.00e-2$\pm$ 7e-5 & 0.878$\pm$ 2e-4 & 0.883$\pm$ 1e-3 & 0.890$\pm$ 2e-3 \\
		&                                      &          1.0          &          1.0           & 2.04e-2$\pm$ 1e-4 & 0.874$\pm$ 4e-4 & 0.845$\pm$ 1e-3 & 0.888$\pm$ 2e-3 \\
		&                                      &          1.0          &          10.0          & 1.87e-2$\pm$ 9e-5 & 0.874$\pm$ 5e-4 & 0.835$\pm$ 1e-3 & 0.885$\pm$ 2e-3 \\
		&                                      &         10.0          &          0.0           & 2.70e-2$\pm$ 9e-5 & 0.812$\pm$ 5e-4 & 0.837$\pm$ 8e-4 & 0.857$\pm$ 4e-3 \\
		&                                      &         10.0          &          0.1           & 2.43e-2$\pm$ 8e-5 & 0.855$\pm$ 3e-4 & 0.838$\pm$ 2e-3 & 0.850$\pm$ 3e-3 \\
		&                                      &         10.0          &          1.0           & 2.33e-2$\pm$ 7e-5 & 0.859$\pm$ 4e-4 & 0.847$\pm$ 1e-3 & 0.848$\pm$ 2e-3 \\
		&                                      &         10.0          &          10.0          & 2.47e-2$\pm$ 9e-5 & 0.835$\pm$ 5e-4 & 0.804$\pm$ 1e-3 & 0.821$\pm$ 2e-3 \\
		\cmidrule{2-8}        & \multirow{8}{*}{$50\%\!\sim\!100\%$} &          1.0          &          0.0           & 2.02e-2$\pm$ 8e-5 & 0.729$\pm$ 3e-4 & 0.706$\pm$ 2e-3 & 0.919$\pm$ 1e-3 \\
		&                                      &          1.0          &          0.1           & 1.90e-2$\pm$ 1e-4 & 0.749$\pm$ 4e-4 & 0.693$\pm$ 7e-4 & 0.913$\pm$ 1e-3 \\
		&                                      &          1.0          &          1.0           & 1.94e-2$\pm$ 1e-4 & 0.752$\pm$ 5e-4 & 0.709$\pm$ 2e-3 & 0.911$\pm$ 9e-4 \\
		&                                      &          1.0          &          10.0          & 1.96e-2$\pm$ 9e-5 & 0.730$\pm$ 4e-4 & 0.642$\pm$ 3e-3 & 0.895$\pm$ 5e-4 \\
		&                                      &         10.0          &          0.0           & 2.88e-2$\pm$ 8e-5 & 0.633$\pm$ 8e-4 & 0.602$\pm$ 3e-3 & 0.869$\pm$ 1e-3 \\
		&                                      &         10.0          &          0.1           & 2.60e-2$\pm$ 5e-5 & 0.671$\pm$ 4e-4 & 0.608$\pm$ 1e-3 & 0.864$\pm$ 8e-4 \\
		&                                      &         10.0          &          1.0           & 2.56e-2$\pm$ 3e-5 & 0.673$\pm$ 6e-4 & 0.595$\pm$ 3e-3 & 0.856$\pm$ 7e-4 \\
		&                                      &         10.0          &          10.0          & 3.01e-2$\pm$ 9e-5 & 0.620$\pm$ 5e-4 & 0.525$\pm$ 3e-3 & 0.820$\pm$ 2e-3 \\ \bottomrule
	\end{tabular}
	\caption{Performance evaluated on $\num[group-separator={,}]{10000}$ images containing $2 \!\sim\! 4$ objects. The models are trained on $\num[group-separator={,}]{50000}$ images containing $2 \!\sim\! 4$ objects using different hyperparameters. The base model is GMIOO.}
	\label{tab:gmioo_compare_test}
\end{table*}

\begin{table*}[t]
	\small
	\centering
	\addtolength{\tabcolsep}{-0pt}
	\begin{tabular}{c|c|c|c|C{0.95in}C{0.95in}C{0.95in}C{0.95in}}
		\toprule
		Dataset            &               Avg Occl               & $\beta_{\text{stg2}}$ & $\alpha_{\text{stg2}}$ &        MSE        &       AMI       &       OCA       &       OOA       \\ \midrule
		\multirow{16.5}{*}{MNIST}   &  \multirow{8}{*}{$0\%\!\sim\!50\%$}  &          1.0          &          0.0           & 1.71e-1$\pm$ 2e-4 & 0.734$\pm$ 6e-4 & 0.469$\pm$ 3e-3 &       N/A       \\
		&                                      &          1.0          &          0.1           & 1.47e-1$\pm$ 2e-4 & 0.761$\pm$ 2e-4 & 0.410$\pm$ 6e-3 &       N/A       \\
		&                                      &          1.0          &          1.0           & 1.29e-1$\pm$ 1e-4 & 0.792$\pm$ 2e-4 & 0.562$\pm$ 4e-3 &       N/A       \\
		&                                      &          1.0          &          10.0          & 1.15e-1$\pm$ 2e-4 & 0.815$\pm$ 4e-4 & 0.566$\pm$ 3e-3 &       N/A       \\
		&                                      &         10.0          &          0.0           & 1.85e-1$\pm$ 1e-4 & 0.699$\pm$ 4e-4 & 0.445$\pm$ 2e-3 &       N/A       \\
		&                                      &         10.0          &          0.1           & 1.93e-1$\pm$ 3e-4 & 0.712$\pm$ 3e-4 & 0.258$\pm$ 4e-3 &       N/A       \\
		&                                      &         10.0          &          1.0           & 1.75e-1$\pm$ 3e-4 & 0.760$\pm$ 4e-4 & 0.542$\pm$ 5e-3 &       N/A       \\
		&                                      &         10.0          &          10.0          & 1.82e-1$\pm$ 3e-4 & 0.764$\pm$ 1e-4 & 0.492$\pm$ 5e-3 &       N/A       \\
		\cmidrule{2-8}        & \multirow{8}{*}{$50\%\!\sim\!100\%$} &          1.0          &          0.0           & 2.65e-1$\pm$ 3e-4 & 0.521$\pm$ 3e-4 & 0.440$\pm$ 2e-3 &       N/A       \\
		&                                      &          1.0          &          0.1           & 2.28e-1$\pm$ 3e-4 & 0.578$\pm$ 4e-4 & 0.427$\pm$ 2e-3 &       N/A       \\
		&                                      &          1.0          &          1.0           & 2.13e-1$\pm$ 2e-4 & 0.630$\pm$ 3e-4 & 0.423$\pm$ 4e-3 &       N/A       \\
		&                                      &          1.0          &          10.0          & 1.78e-1$\pm$ 4e-4 & 0.666$\pm$ 3e-4 & 0.484$\pm$ 3e-3 &       N/A       \\
		&                                      &         10.0          &          0.0           & 2.44e-1$\pm$ 1e-4 & 0.504$\pm$ 4e-4 & 0.294$\pm$ 3e-3 &       N/A       \\
		&                                      &         10.0          &          0.1           & 2.29e-1$\pm$ 4e-4 & 0.539$\pm$ 6e-4 & 0.310$\pm$ 6e-3 &       N/A       \\
		&                                      &         10.0          &          1.0           & 2.30e-1$\pm$ 1e-4 & 0.576$\pm$ 3e-4 & 0.243$\pm$ 3e-3 &       N/A       \\
		&                                      &         10.0          &          10.0          & 2.36e-1$\pm$ 2e-4 & 0.596$\pm$ 5e-4 & 0.317$\pm$ 3e-3 &       N/A       \\ \midrule
		\multirow{16.5}{*}{AbsScene} &  \multirow{8}{*}{$0\%\!\sim\!50\%$}  &          1.0          &          0.0           & 1.68e-2$\pm$ 1e-4 & 0.863$\pm$ 3e-4 & 0.678$\pm$ 3e-3 & 0.866$\pm$ 9e-4 \\
		&                                      &          1.0          &          0.1           & 1.56e-2$\pm$ 1e-4 & 0.869$\pm$ 4e-4 & 0.694$\pm$ 2e-3 & 0.862$\pm$ 1e-3 \\
		&                                      &          1.0          &          1.0           & 1.60e-2$\pm$ 7e-5 & 0.867$\pm$ 3e-4 & 0.661$\pm$ 3e-3 & 0.864$\pm$ 2e-3 \\
		&                                      &          1.0          &          10.0          & 1.58e-2$\pm$ 6e-5 & 0.864$\pm$ 2e-4 & 0.637$\pm$ 1e-3 & 0.857$\pm$ 1e-3 \\
		&                                      &         10.0          &          0.0           & 2.51e-2$\pm$ 7e-5 & 0.807$\pm$ 6e-4 & 0.554$\pm$ 2e-3 & 0.825$\pm$ 3e-3 \\
		&                                      &         10.0          &          0.1           & 2.33e-2$\pm$ 8e-5 & 0.831$\pm$ 4e-4 & 0.547$\pm$ 1e-3 & 0.820$\pm$ 3e-3 \\
		&                                      &         10.0          &          1.0           & 2.25e-2$\pm$ 2e-4 & 0.837$\pm$ 3e-4 & 0.553$\pm$ 2e-3 & 0.822$\pm$ 1e-3 \\
		&                                      &         10.0          &          10.0          & 2.77e-2$\pm$ 6e-5 & 0.787$\pm$ 4e-4 & 0.403$\pm$ 1e-3 & 0.775$\pm$ 2e-3 \\
		\cmidrule{2-8}        & \multirow{8}{*}{$50\%\!\sim\!100\%$} &          1.0          &          0.0           & 2.32e-2$\pm$ 1e-4 & 0.758$\pm$ 5e-4 & 0.436$\pm$ 3e-3 & 0.880$\pm$ 8e-4 \\
		&                                      &          1.0          &          0.1           & 2.22e-2$\pm$ 8e-5 & 0.772$\pm$ 3e-4 & 0.443$\pm$ 2e-3 & 0.873$\pm$ 6e-4 \\
		&                                      &          1.0          &          1.0           & 2.29e-2$\pm$ 7e-5 & 0.772$\pm$ 3e-4 & 0.450$\pm$ 2e-3 & 0.868$\pm$ 6e-4 \\
		&                                      &          1.0          &          10.0          & 2.41e-2$\pm$ 7e-5 & 0.747$\pm$ 5e-4 & 0.386$\pm$ 3e-3 & 0.846$\pm$ 1e-3 \\
		&                                      &         10.0          &          0.0           & 3.43e-2$\pm$ 1e-4 & 0.672$\pm$ 3e-4 & 0.299$\pm$ 2e-3 & 0.824$\pm$ 9e-4 \\
		&                                      &         10.0          &          0.1           & 3.21e-2$\pm$ 8e-5 & 0.698$\pm$ 2e-4 & 0.300$\pm$ 2e-3 & 0.819$\pm$ 1e-3 \\
		&                                      &         10.0          &          1.0           & 3.15e-2$\pm$ 7e-5 & 0.701$\pm$ 5e-4 & 0.316$\pm$ 4e-3 & 0.816$\pm$ 1e-3 \\
		&                                      &         10.0          &          10.0          & 3.70e-2$\pm$ 6e-5 & 0.651$\pm$ 4e-4 & 0.227$\pm$ 2e-3 & 0.779$\pm$ 2e-3 \\ \bottomrule
	\end{tabular}
	\caption{Performance evaluated on $\num[group-separator={,}]{10000}$ images containing $5 \!\sim\! 6$ objects. The models are trained on $\num[group-separator={,}]{50000}$ images containing $2 \!\sim\! 4$ objects using different hyperparameters. The base model is GMIOO.}
	\label{tab:gmioo_compare_general}
\end{table*}

\clearpage

\begin{table*}[t]
	\small
	\centering
	\addtolength{\tabcolsep}{-0pt}
	\begin{tabular}{c|c|c|C{1.0in}C{1.0in}C{1.0in}C{1.0in}}
		\toprule
		Dataset           &               Avg Occl               & Config &            MSE             &           AMI            &           OCA            &           OOA            \\ \midrule
		\multirow{6.5}{*}{MNIST}   &  \multirow{3}{*}{$0\%\!\sim\!50\%$}  & Basic  &            N/A             &     0.716$\pm$ 4e-4      &     0.692$\pm$ 2e-3      &           N/A            \\
		&                                      & Direct &     2.16e-1$\pm$ 5e-4      &     0.711$\pm$ 4e-4      &     0.676$\pm$ 2e-3      &           N/A            \\
		&                                      &  ADI   & \textbf{1.67e-1}$\pm$ 4e-4 & \textbf{0.789}$\pm$ 7e-4 & \textbf{0.739}$\pm$ 4e-3 &           N/A            \\
		\cmidrule{2-7}        & \multirow{3}{*}{$50\%\!\sim\!100\%$} & Basic  &            N/A             &     0.332$\pm$ 4e-4      &     0.186$\pm$ 2e-3      &           N/A            \\
		&                                      & Direct &     3.12e-1$\pm$ 6e-5      &     0.382$\pm$ 5e-4      &     0.482$\pm$ 3e-3      &           N/A            \\
		&                                      &  ADI   & \textbf{2.37e-1}$\pm$ 4e-4 & \textbf{0.452}$\pm$ 3e-4 & \textbf{0.484}$\pm$ 4e-3 &           N/A            \\ \midrule
		\multirow{6.5}{*}{AbsScene} &  \multirow{3}{*}{$0\%\!\sim\!50\%$}  & Basic  &            N/A             &     0.446$\pm$ 3e-4      &     0.492$\pm$ 3e-8      &     0.601$\pm$ 2e-3      \\
		&                                      & Direct &     3.34e-2$\pm$ 2e-4      &     0.795$\pm$ 8e-4      &     0.801$\pm$ 1e-3      &     0.785$\pm$ 4e-3      \\
		&                                      &  ADI   & \textbf{2.58e-2}$\pm$ 7e-5 & \textbf{0.853}$\pm$ 4e-4 & \textbf{0.811}$\pm$ 1e-3 & \textbf{0.858}$\pm$ 3e-3 \\
		\cmidrule{2-7}        & \multirow{3}{*}{$50\%\!\sim\!100\%$} & Basic  &            N/A             &     0.305$\pm$ 3e-4      &     0.157$\pm$ 0e-0      &     0.654$\pm$ 1e-3      \\
		&                                      & Direct &     5.86e-2$\pm$ 7e-5      &     0.358$\pm$ 2e-4      &     0.133$\pm$ 1e-3      &     0.697$\pm$ 1e-3      \\
		&                                      &  ADI   & \textbf{2.52e-2}$\pm$ 2e-4 & \textbf{0.696}$\pm$ 3e-4 & \textbf{0.601}$\pm$ 3e-3 & \textbf{0.885}$\pm$ 1e-3 \\ \bottomrule
	\end{tabular}
	\caption{Performance evaluated on $\num[group-separator={,}]{10000}$ images containing $2 \!\sim\! 4$ objects under different configurations. The models are trained on $\num[group-separator={,}]{50000}$ images containing $2 \!\sim\! 4$ objects. The base model is AIR.}
	\label{tab:air_performance_test}
\end{table*}

\begin{table*}[t]
	\small
	\centering
	\addtolength{\tabcolsep}{-0pt}
	\begin{tabular}{c|c|c|c|C{0.95in}C{0.95in}C{0.95in}C{0.95in}}
		\toprule
		Dataset            &               Avg Occl               & $\beta_{\text{stg2}}$ & $\alpha_{\text{stg2}}$ &        MSE        &       AMI       &       OCA       &       OOA       \\ \midrule
		\multirow{16.5}{*}{MNIST}   &  \multirow{8}{*}{$0\%\!\sim\!50\%$}  &          1.0          &          0.0           & 1.85e-1$\pm$ 3e-4 & 0.767$\pm$ 4e-4 & 0.709$\pm$ 2e-3 &       N/A       \\
		&                                      &          1.0          &          0.1           & 1.78e-1$\pm$ 6e-4 & 0.776$\pm$ 6e-4 & 0.685$\pm$ 3e-3 &       N/A       \\
		&                                      &          1.0          &          1.0           & 1.67e-1$\pm$ 4e-4 & 0.789$\pm$ 7e-4 & 0.739$\pm$ 4e-3 &       N/A       \\
		&                                      &          1.0          &          10.0          & 1.55e-1$\pm$ 3e-4 & 0.792$\pm$ 7e-4 & 0.543$\pm$ 4e-3 &       N/A       \\
		&                                      &         10.0          &          0.0           & 1.89e-1$\pm$ 2e-4 & 0.778$\pm$ 5e-4 & 0.698$\pm$ 2e-3 &       N/A       \\
		&                                      &         10.0          &          0.1           & 1.80e-1$\pm$ 2e-4 & 0.777$\pm$ 3e-4 & 0.608$\pm$ 2e-3 &       N/A       \\
		&                                      &         10.0          &          1.0           & 1.75e-1$\pm$ 5e-4 & 0.791$\pm$ 5e-4 & 0.681$\pm$ 3e-3 &       N/A       \\
		&                                      &         10.0          &          10.0          & 1.77e-1$\pm$ 3e-4 & 0.793$\pm$ 5e-4 & 0.730$\pm$ 3e-3 &       N/A       \\
		\cmidrule{2-8}        & \multirow{8}{*}{$50\%\!\sim\!100\%$} &          1.0          &          0.0           & 2.63e-1$\pm$ 2e-4 & 0.431$\pm$ 5e-4 & 0.531$\pm$ 3e-3 &       N/A       \\
		&                                      &          1.0          &          0.1           & 2.50e-1$\pm$ 2e-4 & 0.437$\pm$ 5e-4 & 0.517$\pm$ 2e-3 &       N/A       \\
		&                                      &          1.0          &          1.0           & 2.37e-1$\pm$ 4e-4 & 0.452$\pm$ 3e-4 & 0.484$\pm$ 4e-3 &       N/A       \\
		&                                      &          1.0          &          10.0          & 2.23e-1$\pm$ 4e-4 & 0.475$\pm$ 5e-4 & 0.343$\pm$ 1e-3 &       N/A       \\
		&                                      &         10.0          &          0.0           & 2.64e-1$\pm$ 4e-4 & 0.385$\pm$ 5e-4 & 0.468$\pm$ 5e-3 &       N/A       \\
		&                                      &         10.0          &          0.1           & 2.49e-1$\pm$ 3e-4 & 0.411$\pm$ 6e-4 & 0.493$\pm$ 3e-3 &       N/A       \\
		&                                      &         10.0          &          1.0           & 2.32e-1$\pm$ 4e-4 & 0.447$\pm$ 8e-4 & 0.496$\pm$ 4e-3 &       N/A       \\
		&                                      &         10.0          &          10.0          & 2.23e-1$\pm$ 2e-4 & 0.456$\pm$ 2e-4 & 0.420$\pm$ 4e-3 &       N/A       \\ \midrule
		\multirow{16.5}{*}{AbsScene} &  \multirow{8}{*}{$0\%\!\sim\!50\%$}  &          1.0          &          0.0           & 2.89e-2$\pm$ 1e-4 & 0.809$\pm$ 4e-4 & 0.785$\pm$ 1e-3 & 0.856$\pm$ 3e-3 \\
		&                                      &          1.0          &          0.1           & 2.69e-2$\pm$ 7e-5 & 0.830$\pm$ 4e-4 & 0.789$\pm$ 1e-3 & 0.859$\pm$ 3e-3 \\
		&                                      &          1.0          &          1.0           & 2.58e-2$\pm$ 7e-5 & 0.853$\pm$ 4e-4 & 0.811$\pm$ 1e-3 & 0.858$\pm$ 3e-3 \\
		&                                      &          1.0          &          10.0          & 2.68e-2$\pm$ 9e-5 & 0.858$\pm$ 2e-4 & 0.828$\pm$ 1e-3 & 0.860$\pm$ 2e-3 \\
		&                                      &         10.0          &          0.0           & 3.53e-2$\pm$ 1e-4 & 0.770$\pm$ 4e-4 & 0.795$\pm$ 2e-3 & 0.797$\pm$ 1e-3 \\
		&                                      &         10.0          &          0.1           & 3.37e-2$\pm$ 1e-4 & 0.786$\pm$ 3e-4 & 0.794$\pm$ 2e-3 & 0.797$\pm$ 1e-3 \\
		&                                      &         10.0          &          1.0           & 3.16e-2$\pm$ 7e-5 & 0.812$\pm$ 7e-4 & 0.770$\pm$ 2e-3 & 0.794$\pm$ 3e-3 \\
		&                                      &         10.0          &          10.0          & 3.13e-2$\pm$ 5e-5 & 0.821$\pm$ 6e-4 & 0.812$\pm$ 2e-3 & 0.801$\pm$ 4e-3 \\
		\cmidrule{2-8}        & \multirow{8}{*}{$50\%\!\sim\!100\%$} &          1.0          &          0.0           & 2.84e-2$\pm$ 7e-5 & 0.645$\pm$ 4e-4 & 0.565$\pm$ 2e-3 & 0.878$\pm$ 9e-4 \\
		&                                      &          1.0          &          0.1           & 2.65e-2$\pm$ 2e-4 & 0.669$\pm$ 4e-4 & 0.588$\pm$ 2e-3 & 0.882$\pm$ 7e-4 \\
		&                                      &          1.0          &          1.0           & 2.52e-2$\pm$ 2e-4 & 0.696$\pm$ 3e-4 & 0.601$\pm$ 3e-3 & 0.885$\pm$ 1e-3 \\
		&                                      &          1.0          &          10.0          & 2.45e-2$\pm$ 6e-5 & 0.700$\pm$ 5e-4 & 0.609$\pm$ 1e-3 & 0.882$\pm$ 7e-4 \\
		&                                      &         10.0          &          0.0           & 3.92e-2$\pm$ 1e-4 & 0.561$\pm$ 5e-4 & 0.486$\pm$ 3e-3 & 0.811$\pm$ 3e-3 \\
		&                                      &         10.0          &          0.1           & 3.76e-2$\pm$ 6e-5 & 0.577$\pm$ 2e-4 & 0.502$\pm$ 9e-4 & 0.815$\pm$ 1e-3 \\
		&                                      &         10.0          &          1.0           & 3.46e-2$\pm$ 6e-5 & 0.605$\pm$ 2e-4 & 0.522$\pm$ 4e-3 & 0.821$\pm$ 1e-3 \\
		&                                      &         10.0          &          10.0          & 3.57e-2$\pm$ 1e-4 & 0.590$\pm$ 8e-4 & 0.520$\pm$ 2e-3 & 0.811$\pm$ 1e-3 \\ \bottomrule
	\end{tabular}
	\caption{Performance evaluated on $\num[group-separator={,}]{10000}$ images containing $2 \!\sim\! 4$ objects. The models are trained on $\num[group-separator={,}]{50000}$ images containing $2 \!\sim\! 4$ objects using different hyperparameters. The base model is AIR.}
	\label{tab:air_compare_test}
\end{table*}

\begin{table*}[t]
	\small
	\centering
	\addtolength{\tabcolsep}{-0pt}
	\begin{tabular}{c|c|c|C{1.0in}C{1.0in}C{1.0in}C{1.0in}}
		\toprule
		Dataset           &               Avg Occl               & Config &            MSE             &           AMI            &           OCA            &           OOA            \\ \midrule
		\multirow{6.5}{*}{MNIST}   &  \multirow{3}{*}{$0\%\!\sim\!50\%$}  & Basic  &            N/A             &     0.660$\pm$ 2e-4      &     0.237$\pm$ 2e-3      &           N/A            \\
		&                                      & Direct &     2.45e-1$\pm$ 2e-4      &     0.668$\pm$ 3e-4      &     0.376$\pm$ 2e-3      &           N/A            \\
		&                                      &  ADI   & \textbf{1.79e-1}$\pm$ 3e-4 & \textbf{0.749}$\pm$ 8e-5 & \textbf{0.553}$\pm$ 3e-3 &           N/A            \\
		\cmidrule{2-7}        & \multirow{3}{*}{$50\%\!\sim\!100\%$} & Basic  &            N/A             &     0.478$\pm$ 3e-4      &     0.044$\pm$ 2e-3      &           N/A            \\
		&                                      & Direct &     3.28e-1$\pm$ 4e-4      &     0.514$\pm$ 3e-4      &     0.190$\pm$ 2e-3      &           N/A            \\
		&                                      &  ADI   & \textbf{2.51e-1}$\pm$ 3e-4 & \textbf{0.584}$\pm$ 3e-4 & \textbf{0.413}$\pm$ 3e-3 &           N/A            \\ \midrule
		\multirow{6.5}{*}{AbsScene} &  \multirow{3}{*}{$0\%\!\sim\!50\%$}  & Basic  &            N/A             &     0.352$\pm$ 2e-4      &     0.000$\pm$ 0e-0      &     0.545$\pm$ 3e-4      \\
		&                                      & Direct &     3.30e-2$\pm$ 9e-5      &     0.785$\pm$ 4e-4      &     0.451$\pm$ 3e-3      &     0.753$\pm$ 3e-3      \\
		&                                      &  ADI   & \textbf{2.14e-2}$\pm$ 7e-5 & \textbf{0.843}$\pm$ 7e-5 & \textbf{0.587}$\pm$ 1e-3 & \textbf{0.828}$\pm$ 1e-3 \\
		\cmidrule{2-7}        & \multirow{3}{*}{$50\%\!\sim\!100\%$} & Basic  &            N/A             &     0.311$\pm$ 2e-4      &     0.000$\pm$ 0e-0      &     0.582$\pm$ 1e-3      \\
		&                                      & Direct &     6.55e-2$\pm$ 3e-5      &     0.411$\pm$ 2e-4      &     0.005$\pm$ 3e-4      &     0.655$\pm$ 1e-3      \\
		&                                      &  ADI   & \textbf{2.85e-2}$\pm$ 7e-5 & \textbf{0.731}$\pm$ 2e-4 & \textbf{0.410}$\pm$ 4e-3 & \textbf{0.842}$\pm$ 1e-3 \\ \bottomrule
	\end{tabular}
	\caption{Performance evaluated on $\num[group-separator={,}]{10000}$ images containing $5 \!\sim\! 6$ objects under different configurations. The models are trained on $\num[group-separator={,}]{50000}$ images containing $2 \!\sim\! 4$ objects. The base model is AIR.}
	\label{tab:air_performance_general}
\end{table*}

\begin{table*}[t]
	\small
	\centering
	\addtolength{\tabcolsep}{-0pt}
	\begin{tabular}{c|c|c|c|C{0.95in}C{0.95in}C{0.95in}C{0.95in}}
		\toprule
		Dataset            &               Avg Occl               & $\beta_{\text{stg2}}$ & $\alpha_{\text{stg2}}$ &        MSE        &       AMI       &       OCA       &       OOA       \\ \midrule
		\multirow{16.5}{*}{MNIST}   &  \multirow{8}{*}{$0\%\!\sim\!50\%$}  &          1.0          &          0.0           & 2.04e-1$\pm$ 2e-4 & 0.725$\pm$ 3e-4 & 0.522$\pm$ 3e-3 &       N/A       \\
		&                                      &          1.0          &          0.1           & 1.86e-1$\pm$ 3e-4 & 0.738$\pm$ 1e-4 & 0.506$\pm$ 3e-3 &       N/A       \\
		&                                      &          1.0          &          1.0           & 1.79e-1$\pm$ 3e-4 & 0.749$\pm$ 8e-5 & 0.553$\pm$ 3e-3 &       N/A       \\
		&                                      &          1.0          &          10.0          & 1.63e-1$\pm$ 1e-4 & 0.765$\pm$ 3e-4 & 0.307$\pm$ 3e-3 &       N/A       \\
		&                                      &         10.0          &          0.0           & 2.04e-1$\pm$ 2e-4 & 0.737$\pm$ 5e-4 & 0.530$\pm$ 4e-3 &       N/A       \\
		&                                      &         10.0          &          0.1           & 1.95e-1$\pm$ 2e-4 & 0.742$\pm$ 3e-4 & 0.485$\pm$ 4e-3 &       N/A       \\
		&                                      &         10.0          &          1.0           & 1.86e-1$\pm$ 3e-4 & 0.753$\pm$ 2e-4 & 0.554$\pm$ 5e-3 &       N/A       \\
		&                                      &         10.0          &          10.0          & 1.93e-1$\pm$ 3e-4 & 0.748$\pm$ 7e-4 & 0.531$\pm$ 4e-3 &       N/A       \\
		\cmidrule{2-8}        & \multirow{8}{*}{$50\%\!\sim\!100\%$} &          1.0          &          0.0           & 2.78e-1$\pm$ 4e-4 & 0.561$\pm$ 5e-4 & 0.365$\pm$ 2e-3 &       N/A       \\
		&                                      &          1.0          &          0.1           & 2.66e-1$\pm$ 3e-4 & 0.566$\pm$ 3e-4 & 0.398$\pm$ 2e-3 &       N/A       \\
		&                                      &          1.0          &          1.0           & 2.51e-1$\pm$ 3e-4 & 0.584$\pm$ 3e-4 & 0.413$\pm$ 3e-3 &       N/A       \\
		&                                      &          1.0          &          10.0          & 2.30e-1$\pm$ 3e-4 & 0.611$\pm$ 5e-4 & 0.340$\pm$ 3e-3 &       N/A       \\
		&                                      &         10.0          &          0.0           & 2.80e-1$\pm$ 3e-4 & 0.516$\pm$ 3e-4 & 0.228$\pm$ 2e-3 &       N/A       \\
		&                                      &         10.0          &          0.1           & 2.64e-1$\pm$ 2e-4 & 0.543$\pm$ 2e-4 & 0.324$\pm$ 5e-3 &       N/A       \\
		&                                      &         10.0          &          1.0           & 2.45e-1$\pm$ 3e-4 & 0.579$\pm$ 2e-4 & 0.409$\pm$ 4e-3 &       N/A       \\
		&                                      &         10.0          &          10.0          & 2.32e-1$\pm$ 3e-4 & 0.590$\pm$ 4e-4 & 0.366$\pm$ 4e-3 &       N/A       \\ \midrule
		\multirow{16.5}{*}{AbsScene} &  \multirow{8}{*}{$0\%\!\sim\!50\%$}  &          1.0          &          0.0           & 2.44e-2$\pm$ 2e-4 & 0.821$\pm$ 3e-4 & 0.556$\pm$ 2e-3 & 0.823$\pm$ 2e-3 \\
		&                                      &          1.0          &          0.1           & 2.21e-2$\pm$ 6e-5 & 0.836$\pm$ 1e-4 & 0.571$\pm$ 2e-3 & 0.829$\pm$ 1e-3 \\
		&                                      &          1.0          &          1.0           & 2.14e-2$\pm$ 7e-5 & 0.843$\pm$ 7e-5 & 0.587$\pm$ 1e-3 & 0.828$\pm$ 1e-3 \\
		&                                      &          1.0          &          10.0          & 2.28e-2$\pm$ 6e-5 & 0.843$\pm$ 2e-4 & 0.585$\pm$ 3e-3 & 0.819$\pm$ 3e-3 \\
		&                                      &         10.0          &          0.0           & 3.40e-2$\pm$ 5e-5 & 0.768$\pm$ 2e-4 & 0.505$\pm$ 2e-3 & 0.764$\pm$ 1e-3 \\
		&                                      &         10.0          &          0.1           & 3.33e-2$\pm$ 9e-5 & 0.773$\pm$ 6e-4 & 0.493$\pm$ 3e-3 & 0.760$\pm$ 1e-3 \\
		&                                      &         10.0          &          1.0           & 3.10e-2$\pm$ 1e-4 & 0.789$\pm$ 2e-4 & 0.492$\pm$ 2e-3 & 0.764$\pm$ 3e-3 \\
		&                                      &         10.0          &          10.0          & 3.05e-2$\pm$ 5e-5 & 0.799$\pm$ 9e-5 & 0.532$\pm$ 8e-4 & 0.772$\pm$ 1e-3 \\
		\cmidrule{2-8}        & \multirow{8}{*}{$50\%\!\sim\!100\%$} &          1.0          &          0.0           & 3.11e-2$\pm$ 7e-5 & 0.699$\pm$ 4e-4 & 0.395$\pm$ 3e-3 & 0.836$\pm$ 9e-4 \\
		&                                      &          1.0          &          0.1           & 2.96e-2$\pm$ 8e-5 & 0.715$\pm$ 1e-4 & 0.396$\pm$ 4e-3 & 0.838$\pm$ 1e-3 \\
		&                                      &          1.0          &          1.0           & 2.85e-2$\pm$ 7e-5 & 0.731$\pm$ 2e-4 & 0.410$\pm$ 4e-3 & 0.842$\pm$ 1e-3 \\
		&                                      &          1.0          &          10.0          & 2.90e-2$\pm$ 3e-5 & 0.728$\pm$ 5e-4 & 0.417$\pm$ 4e-3 & 0.836$\pm$ 1e-3 \\
		&                                      &         10.0          &          0.0           & 4.37e-2$\pm$ 6e-5 & 0.623$\pm$ 2e-4 & 0.314$\pm$ 2e-3 & 0.781$\pm$ 1e-3 \\
		&                                      &         10.0          &          0.1           & 4.25e-2$\pm$ 7e-5 & 0.637$\pm$ 2e-4 & 0.308$\pm$ 4e-3 & 0.784$\pm$ 1e-3 \\
		&                                      &         10.0          &          1.0           & 3.93e-2$\pm$ 7e-5 & 0.662$\pm$ 5e-4 & 0.353$\pm$ 2e-3 & 0.791$\pm$ 1e-3 \\
		&                                      &         10.0          &          10.0          & 4.13e-2$\pm$ 7e-5 & 0.644$\pm$ 3e-4 & 0.326$\pm$ 3e-3 & 0.779$\pm$ 7e-4 \\ \bottomrule
	\end{tabular}
	\caption{Performance evaluated on $\num[group-separator={,}]{10000}$ images containing $5 \!\sim\! 6$ objects. The models are trained on $\num[group-separator={,}]{50000}$ images containing $2 \!\sim\! 4$ objects using different hyperparameters. The base model is AIR.}
	\label{tab:air_compare_general}
\end{table*}

\end{document}